\documentclass[10pt,twocolumn,letterpaper]{article}

\usepackage{cvpr}              % To produce the CAMERA-READY version
\usepackage{graphicx}
\usepackage{amsmath}
\usepackage{amssymb}
\usepackage{booktabs}
\usepackage{float}
\usepackage{multirow}
\usepackage{subcaption}
\usepackage[super]{nth}
\usepackage[accsupp]{axessibility}  % Improves PDF readability for those with disabilities.

\usepackage[pagebackref,breaklinks,colorlinks]{hyperref}

\usepackage[capitalize]{cleveref}
\crefname{section}{Sec.}{Secs.}
\Crefname{section}{Section}{Sections}
\Crefname{table}{Table}{Tables}
\crefname{table}{Tab.}{Tabs.}

\def\cvprPaperID{6512} % *** Enter the CVPR Paper ID here
\def\confName{CVPR}
\def\confYear{2022}

\begin{document}

\title{Practical Stereo Matching via Cascaded Recurrent Network \linebreak with Adaptive Correlation}

\author{
    Jiankun Li$^{1}$\ \ \  Peisen Wang$^{1}$\footnotemark[1] \ \ \ Pengfei Xiong$^{2}$\footnotemark[1] \ \ \ Tao Cai$^{1}$\ \ \ Ziwei Yan$^{1}$\ \ \ Lei Yang$^{1}$\ \ \  
    \\
    Jiangyu Liu$^{1}$\ \ \ Haoqiang Fan$^{1}$\ \ \ Shuaicheng Liu$^{3,1}$\footnotemark[2]
    \\
    $^{1}$Megvii Research \quad  $^{2}$Tencent \\
    $^{3}$University of Electronic Science and Technology of China \\
    \url{https://github.com/megvii-research/CREStereo}
    \vspace{-0.5em}
}

% \maketitle

\twocolumn[{
\renewcommand\twocolumn[1][]{#1}
\maketitle
\begin{center}
    \centering
    \vspace{-0.5em}
    \includegraphics[width=0.96\linewidth,page=1]{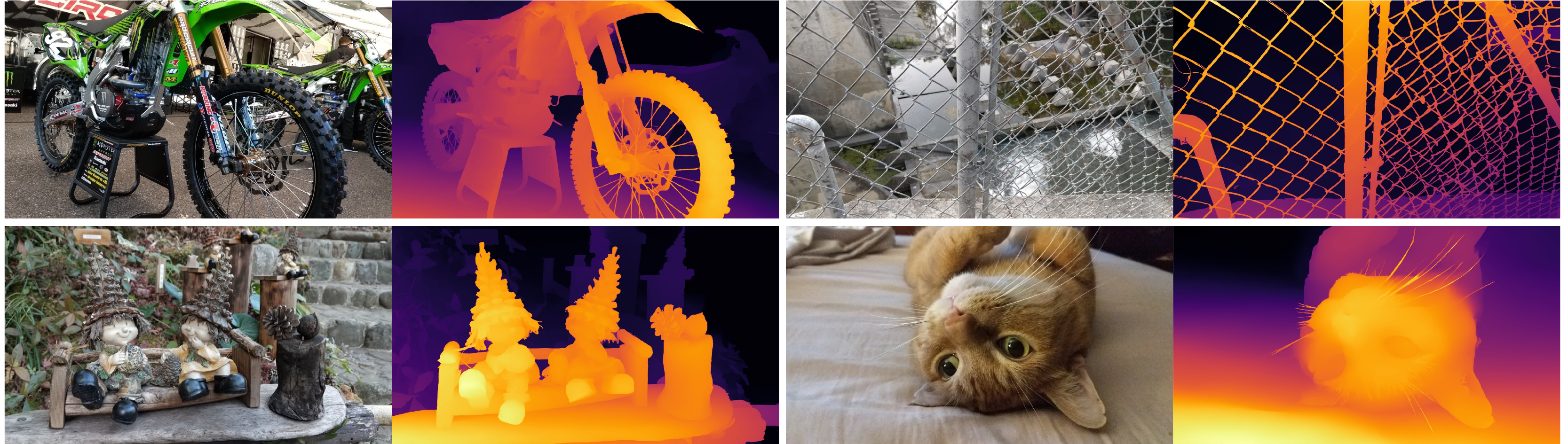}
    \vspace{-0.5em}
    \captionof{figure}{Examples of our predictions on images from Holopix50K  \cite{hua2020holopix50k} dataset. We show left images of the stereo pairs and their corresponding predicted disparities. Our results achieve high accuracy and exhibit high-quality details for fine-structured objects.}
    \label{fig1:our_result}
    \vspace{-0.5em}
\end{center}
}]

%%%%%%%%% ABSTRACT
\begin{abstract} 
\vspace{-1em}
With the advent of convolutional neural networks, stereo matching algorithms have recently gained tremendous progress. 
However, it remains a great challenge to accurately extract disparities from real-world image pairs taken by consumer-level devices like smartphones, due to practical complicating factors such as thin structures, non-ideal rectification, camera module inconsistencies and various hard-case scenes. 
In this paper, we propose a set of innovative designs to tackle the problem of practical stereo matching: 1) to better recover fine depth details, we design a hierarchical network with recurrent refinement to update disparities in a coarse-to-fine manner, as well as a stacked cascaded architecture for inference; 2) we propose an adaptive group correlation layer to mitigate the impact of erroneous rectification; 3) we introduce a new synthetic dataset with special attention to difficult cases for better generalizing to real-world scenes. Our results not only rank \nth{1} on both Middlebury and ETH3D benchmarks, outperforming existing state-of-the-art methods by a notable margin, but also exhibit high-quality details for real-life photos, which clearly demonstrates the efficacy of our contributions. 
% Datasets and codes are available at \url{https://github.com/megvii-research/CREStereo}.
\end{abstract}

\renewcommand{\thefootnote}{\fnsymbol{footnote}}
\newcounter{somecounter}
\setcounter{somecounter}{2}
\footnotetext[1]{Equal contribution. \fnsymbol{somecounter} Corresponding author.}
% \footnotetext[2]{Corresponding author.}
\renewcommand{\thefootnote}{\arabic{footnote}}

%%%%%%%%% BODY TEXT
\section{Introduction}

Stereo matching is a classical research topic of computer vision, the goal of which, 
given a pair of rectified images, is to compute the displacement between two corresponding pixels, namely ``disparity" \cite{scharstein2002taxonomy}. It plays an important role in many applications, including autonomous driving, augmented reality, simulated bokeh rendering and so forth.

Recently, with the support of large synthetic datasets \cite{mayer2016large, sintel, fallingthings}, convolutional neural network (CNN) based stereo matching methods have taken the accuracy of disparity estimation to a new height \cite{lipson2021raft, cheng2020hierarchical, tankovich2021hitnet}. However, to make the algorithm truly practical in the scenario of everyday consumer photography, we are still faced with three major obstacles. 

Firstly, it remains a complicated issue for most existing algorithms to precisely recover the disparity of fine image details, or thin structures such as nets and wire frames. The fact that consumer photos are being produced in higher resolutions only serves to worsen the problem. In computational bokeh, for instance, disparity error around fine details would result in degraded rendering results that are unpleasing to human perception \cite{bokeh}. Secondly, perfect rectification \cite{zhang1998determining, loop1999computing} is hard to obtain for real-world stereo image pairs, as they are often produced by camera modules with different traits. For example, most current smartphones capture the stereo pair with a wide-angle and a telephoto lens, which have distinct characteristics like focal length and distortion parameters, inevitably resulting in non-ideal rectifications. Therefore existing methods assuming that the stereo pair is perfectly rectified are likely to fail under such adversarial conditions. In addition, the image pair produced by inconsistent cameras modules may vary in illumination, white balancing, image quality, etc., making the estimation task even harder. Finally, though it has been shown that models trained from large enough synthetic datasets can generalize well to real-world scenes\cite{flownet, mayer2016large}, disparity estimation in typical hard cases, like non-texture or repetitive-texture regions, continues to be difficult, which requires special attention be paid in covering relevant scenes in the training dataset.

In this paper, we propose CREStereo, namely Cascaded REcurrent Stereo matching network, which comprises a set of novel designs, to tackle the problem of practical stereo matching. To better recover intricate image details, we design a hierarchical network to update disparities recurrently in a coarse-to-fine manner; in addition, we adopt a stacked cascaded architecture for high-resolution inference. To alleviate the negative influence of rectification error, we design an adaptive group local correlation layer for feature matching. Furthermore, we introduce a new synthetic dataset with richer variations in lighting, texture and shapes, in order to better generalize to real-world scenes.

So far, CREStereo ranks \nth{1} on both ETH3D two-view stereo \cite{eth3d} and Middlebury \cite{middlebury} benchmarks, and achieves competitive performance on KITTI 2012/2015 \cite{kitti} among published methods. Additionally, our network demonstrates superior performance for arbitrary real-world scenes, well proving the effectiveness of our designs.

Our main contributions can thus be summarized as follows: 1) we propose a cascaded recurrent network for practical stereo matching and a stacked cascaded architecture for high-resolution inference; 2) we design an adaptive group correlation layer to handle non-ideal rectification; 3) we create a new synthetic dataset to better generalize to real-world scenes; 4) our method outperforms existing methods on public benchmarks such as Middlebury and ETH3D by a significant margin, and considerably increases the accuracy of recovered disparities for real-world stereo images.

\begin{figure*}[ht]
    \centering
    \includegraphics[width=1\linewidth,page=1]{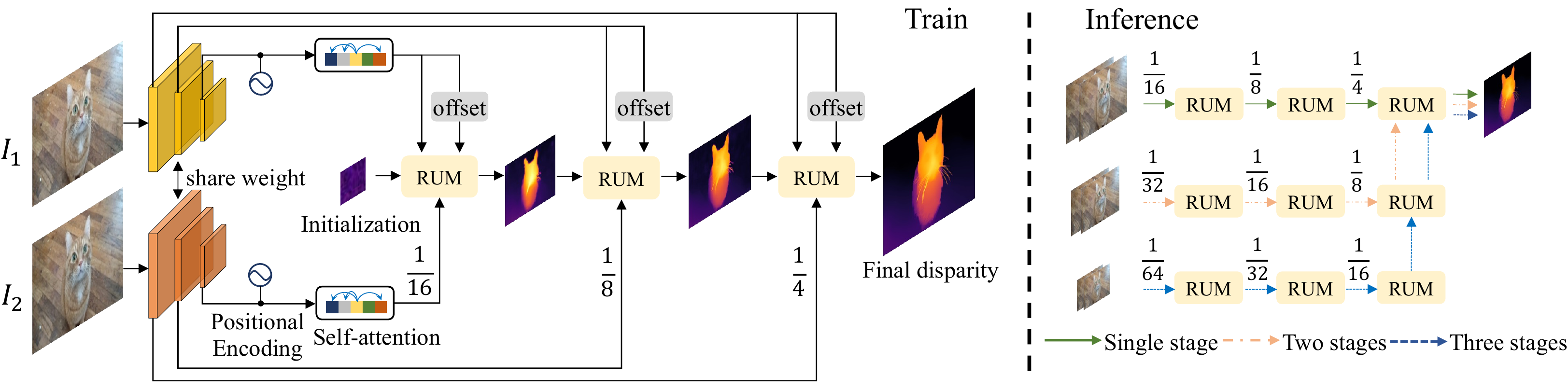}
    \vspace{-1.5em}
    \caption{An overview of our proposed network. Left: A pair of stereo images $I_{1}$ and $I_{2}$ are fed into two shared-weight feature extraction networks to produce a 3-level feature pyramid, which is used to compute different scales of correlations in the 3 stages of cascaded recurrent networks. The feature pyramid of $I_{1}$ also provides context information for latter update blocks and offsets computation. In each stage of the cascades, the features and the predicted disparities are refined iteratively using the Recurrent Update Module (RUM, Sec.~\ref{CRN}), and the final output disparity of the former stage is fed to the next as an initialization. For each iteration in RUM, we apply Adaptive Group Correlation Layer (AGCL, Sec.~\ref{AGCL}) to compute the correlation. Right: Our proposed stacked cascaded architecture in inference phase, which takes an image pyramid as input, taking advantage of multi-level context, as detailed in Sec.~\ref{Infer} . }
    \label{fig2:Architecture}
    \vspace{-0.2cm}
\end{figure*}

%-------------------------------------------------------------------------
\section{Related Work}

\textbf{Traditional algorithms.}
Stereo matching is a challenging problem and has been studied for a long time. Traditional algorithms can be categorized into local and global methods. Local methods \cite{birchfield1999depth, hirschmuller2002real, van2002hierarchical} compute matching cost using a support window centered at pixels along the epipolar line. Global methods treat stereo matching as an optimization problem, where an explicit cost function is formulated and optimized by \textit{belief propagation} \cite{sun2003stereo, klaus2006segment, yang2008stereo} or \textit{graph cut} \cite{boykov2001fast} algorithms. A semi-global matching (SGM) method is later proposed \cite{hirschmuller2005accurate} using mutual information instead of intensity based on dynamic programming.

\textbf{Learning-based algorithms.}
Deep neural network was first introduced in stereo matching task only for matching cost computation. Zbontar and LeCun \cite{zbontar2015computing} proposed to train a CNN to initialize the matching cost between patches, which is refined by cross-based aggregation and semi-global optimization as in SGM \cite{hirschmuller2005accurate}. In recent years, end-to-end network has become mainstream in stereo matching. One line of networks \cite{mayer2016large, pang2017cascade, liang2018learning, guo2019group, xu2020aanet, tankovich2021hitnet, li2020revisiting} only uses 2D convolutions. Mayer \etal \cite{mayer2016large} introduced the first end-to-end network named DispNet and its correlation version DispNetC for disparity estimation. Pang \etal \cite{pang2017cascade} proposed a two-stage framework called CRL with multi-scale residual learning.  Guo \etal \cite{guo2019group} proposed GwcNet with group-wise correlation to improve similarity measurement. AANet \cite{xu2020aanet} introduced a novel aggregation method using sparse points and multi-scale interaction.  A very recent method, RAFT-Stereo \cite{lipson2021raft}, takes advantage of the iterative refinement in the optical flow network RAFT \cite{teed2020raft} to design a network adapted for stereo matching. Another line of networks \cite{kendall2017end, chang2018pyramid, zhang2019ga, khamis2018stereonet, yang2019hierarchical} uses 3D convolutions to perform cost volume construction and aggregation in traditional methods. GCNet\cite{kendall2017end} and PSMNet \cite{chang2018pyramid} proposed to construct a 4D cost volume with 3D hourglass aggregation networks. For high-res images, Yang \etal \cite{yang2019hierarchical} proposed a coarse-to-fine hierarchical network to address memory and speed issues. Lately, neural architecture search has also been introduced into deep stereo networks \cite{cheng2020hierarchical}.

\textbf{Practical stereo matching.}
Stereo matching oriented toward real-world images is a less explored problem. Pang \etal \cite{pang2018zoom}  proposed a self-adaptation approach for generalizing CNN to target domain without ground truth disparity. Luo \etal \cite{luo2020wavelet} proposed a wavelet synthesis network to produce better results for bokeh applications on smartphones. Song \etal  \cite{song2021adastereo} introduced a domain adaptation pipeline for networks to narrow down the gap between synthetic and real-world domains.

\textbf{Synthetic datasets.} 
Sufficient training data is essential for deep stereo models, but it is hard to obtain accurate disparity in real world. Synthetic datasets 
\cite{mayer2016large, sintel, fallingthings, mayer2018makes} provide high-accuracy and dense ground truth. Recently, He \etal \cite{he2021semi} built a data generation
pipeline for stereo matching using Blender \cite{blender}, with textures from real images of
common datasets. Autoflow \cite{autoflow} introduced a simple method to render
randomized polygons with motion for optical flow training.  Despite the
effectiveness of these datasets, they still have limited variations of object shapes, and a restricted distribution of the disparity/optical flow
values, which weakens the generalizing ability from synthetic to real world.

%------------------------------------------------------------------------

\section{Methods}

\begin{figure*}[ht]
    \centering
    \includegraphics[width=1\linewidth,page=1]{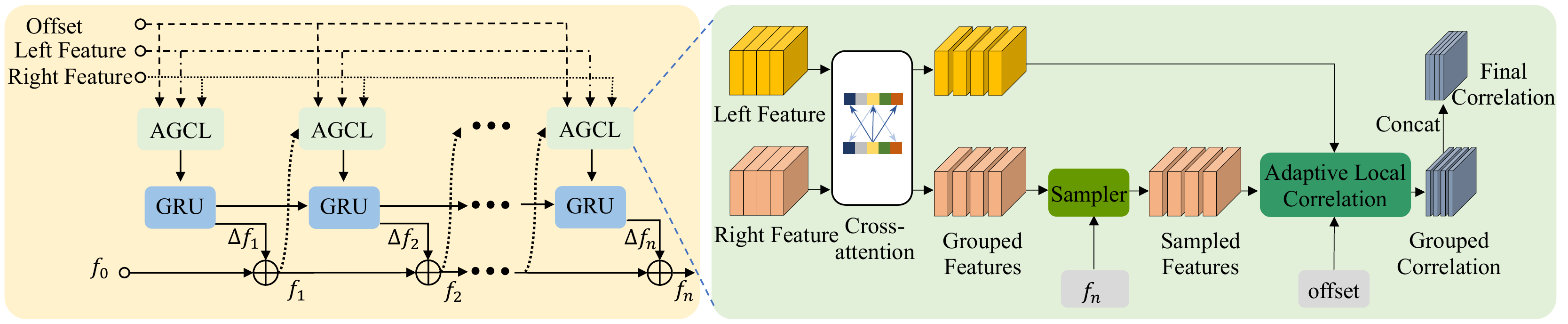}
    \vspace{-1.5em}
    \caption{The architecture of proposed modules. Left: Recurrent Update Module (RUM). Right: Adaptive Group Correlation Layer (AGCL). Details are described in Sec.~\ref{CRN}  and Sec.~\ref{AGCL}, respectively.}
    \label{fig3:Architecture2}
    \vspace{-0.2cm}
\end{figure*}

In this section we present the key components of the proposed Cascaded REcurrent Stereo matching network (CREStereo) and our new synthetic dataset.

\subsection{Adaptive Group Correlation Layer \label{AGCL}}

We observe that it is difficult to implement perfect calibration for real-world stereo cameras. For instance, the two cameras may not be strictly placed on the horizontal epipolar line, resulting in slight rotations in 3D space; or images from camera lenses usually have residual distortion even after they are rectified. As a result, for a stereo image pair, the corresponding points may not locate on the same scan-line. We thus propose an Adaptive Group Correlation Layer (AGCL) to reduce the matching ambiguity in this situation, achieving better performance compared to all-pairs matching \cite{teed2020raft, lipson2021raft} while only local correlation is computed.

\textbf{Local Feature Attention. }
Instead of computing global correlation for every pair of pixels, we only match points in a local window to avoid large memory consumption and computation cost. In light of LoFTR \cite{sun2021loftr} for sparse feature matching,  we add an attention module before correlation computation in the first stage of cascades in order to aggregate global context information in single or cross feature maps. 
Following \cite{sun2021loftr}, we add positional encoding to the backbone output, which enhances positional dependence of the feature maps. The self and cross attention is computed alternately, where a linear attention layer is used to reduce computation complexity.  

\textbf{2D-1D Alternate Local Search. }
Different from the flow estimation network RAFT \cite{teed2020raft} and its stereo version \cite{lipson2021raft} where all-pairs correlation is computed by a matrix multiplication of two $C \times H \times W$ feature maps, which outputs a 4D $H\times W\times H\times W$ or 3D $H\times W\times W$ cost volume, we only compute correlation in a local search window that outputs a much smaller volume of $H\times W\times D$ to save the memory and computation cost. $H$ and $W$ denote the height and width of the feature maps, and $D$ is the number of correlation pairs much smaller than $W$.
Our correlation computation is also distinct from cost volume based stereo networks like \cite{chang2018pyramid,khamis2018stereonet, yang2019hierarchical, xu2020aanet} where the search range is related to the maximum displacement of foreground objects. This fixed range is much larger than the number of local correlation pairs we use,  which leads to more noisy interference. Furthermore, we don't need to preset the range when the model generalizes to stereo pairs with different baselines.

Given two resampled and attended feature maps $\mathbf{F}_{1}$ and $\mathbf{F}_{2}$, the local correlation at position $(x, y)$ can be denoted as
\begin{equation}
    \mathrm{Corr}(x, y, d) = \frac{1}{C} \sum_{i=1}^{C} \mathbf{F}_{1}(i, x, y) \mathbf{F}_{2}(i, x', y'),
    \label{corr}
\end{equation} where $x' = x + f(d)$, $y' = y + g(d)$, $\mathrm{Corr}(x, y, d) \in \mathbb{R}^{H\times W\times D}$ is the matching cost of $d$-th ($d \in [0, D-1]$) correlation pair,  $C$ is the number of feature channels, $f(d)$ and $g(d)$ denote the fixed offset of current pixels in horizontal  and vertical directions.

Traditionally, search direction between two rectified images only lies on the epipolar line in stereo matching. To deal with non-ideal stereo rectification cases, we adopt a 2D-1D alternate local search strategy to improve the matching accuracy.  
In 1D search mode, we set $g(d)=0$ and $f(d) \in [-r, r]$, where $r=4$. Positive displacement value of $f(d)$ is reserved to adjust inaccurate results after every iterative sampling. The results computed by Eq.~\ref{corr} are stacked and concatenated at channel dimension for the final correlation volume.
In 2D search mode, a $k\times k$ grid with dilation $l$ similar to dilated convolution\cite{yu2017dilated}  is used for correlation computation. We set $k=\sqrt{2r+1}$ to make sure the output features have the same number of channels so that they can be fed to a shared-weight update block. Cooperating with iterative resampling, alternate local search also acts as a propagation module for recurrent refinement, where the network learns to replace the biased prediction on current location with its more accurate neighbors.

\begin{figure}[t]
    \centering
    \includegraphics[width=0.7\linewidth,page=1]{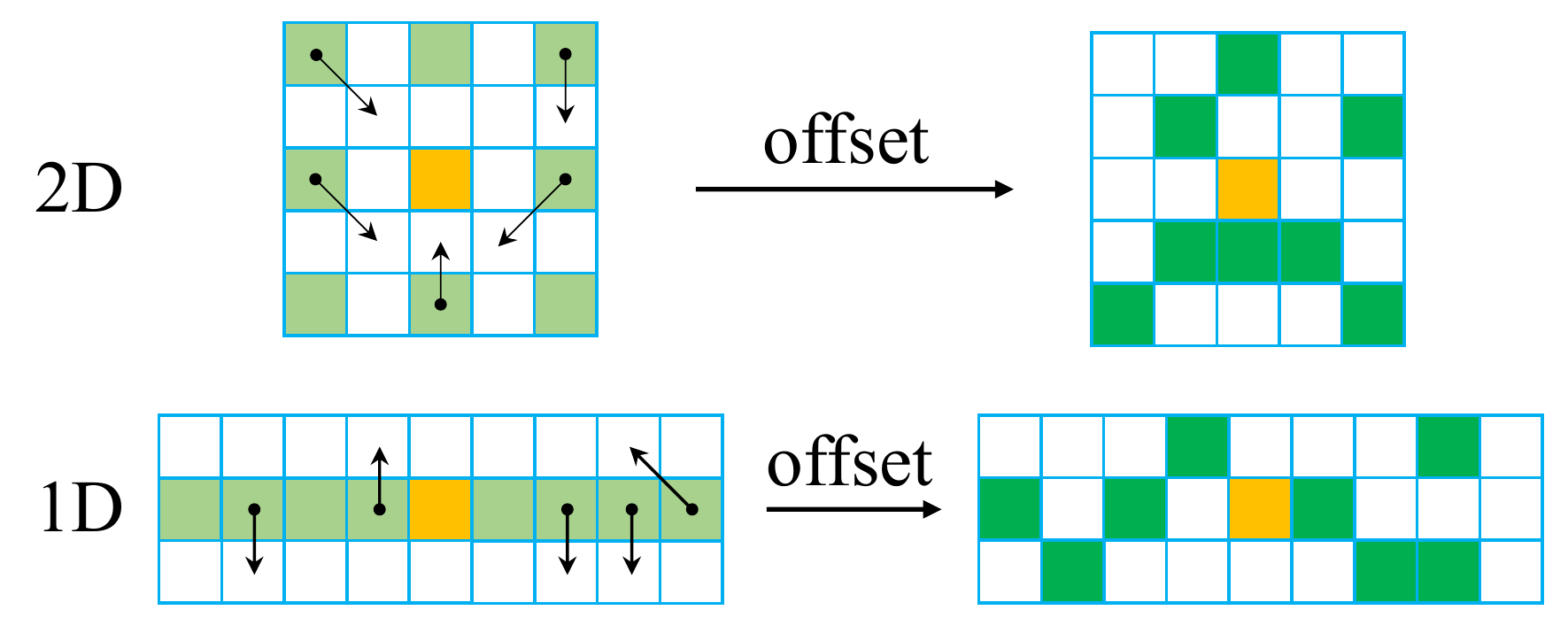}
    \vspace{-0.5em}
    \caption{Illustration of the adaptive local correlation. The top and the bottom are 2D and 1D situations respectively, which share the same number of searched neighbors to produce correlation maps in the same shape.}
    \label{fig2:offset}
    % \vspace{-0.3cm}
\end{figure}

\textbf{Deformable search window.}
 Stereo matching often suffers from ambiguity in occlusion or textureless areas.  Correlation computed in a fixed-shape local search window tends to be vulnerable to those cases. Extending deformable convolution\cite{zhu2019deformable} to correlation computation, we use a content adaptive search window for correlation pairs generation, which is different from AANet\cite{xu2020aanet} where a similar strategy is adopted only in cost aggregation. With the learned additional offset $\mathrm{d}x$ and $\mathrm{d}y$, the new correlation can be computed as
\begin{equation}
  \mathrm{Corr}(x, y, d) = \frac{1}{C} \sum_{i=1}^{C} \mathbf{F}_{1}(i, x, y) \mathbf{F}_{2}(i, x'', y'')
  \label{corr_offset}
\end{equation} where $x'' = x + f(d) + \mathrm{d}x$, $y'' = y + g(d) + \mathrm{d}y$.
Fig.~\ref{fig2:offset} shows how offsets change the formation of a conventional search window. 

\textbf{Group-wise correlation.}
Inspired by \cite{guo2019group} which introduces the group-wise 4D cost volume, we split the feature map into $\mathcal{G}$ groups to compute local correlation individually. Finally, we concatenate $\mathcal{G}$ correlation volumes of $D\times H\times W$ in channel dimension to get the $\mathcal{G}D\times H\times W$ output volume. The procedure is shown in Fig.~\ref{fig3:Architecture2}.

\subsection{Cascaded Recurrent Network \label{CRN}}

For non-texture or repetitive-texture areas, matching is more robust using low-res and high-level feature maps due to large receptive field and sufficient semantic information. However the details of fine structures may be lost in such feature maps. In order to maintain robustness and preserve the details in high-res input simultaneously, we propose cascaded recurrent refinement for correlation computing and disparity updating.

\textbf{Recurrent Update Module. } We build a Recurrent Update Module (RUM) based on GRU blocks and our Adaptive Group Correlation Layer (AGCL). 
Unlike in RAFT\cite{teed2020raft} where the feature pyramid is constructed in a single correlation layer with the output being merged into one volume, we compute correlations for every feature map respectively in different cascade levels and refine the disparities for several iterations independently. As is shown in Fig.~\ref{fig3:Architecture2}, the ``sampler" samples locations of grouped feature taking coordinate grid derived from $f_n$  as input. $\{f_1,...,f_n\}$ are intermediate predictions of $n$ iterations with initialization $f_0$. Current correlation volume is constructed with learned offsets $ o \in \mathbb{R}^{2\times (2r+1) \times h \times w} $. The GRU blocks update current prediction and feed it to the AGCL in next iteration.

\textbf{Cascaded Refinement. } Except for the first level of cascades, which starts at 1/16 of the input resolution with disparity initialized to all zeros, other levels take the upsampled version of prediction from previous level as initialization. Though handling different levels of refinement, all RUMs share the same weights. After the last refinement level, convex upsampling \cite{teed2020raft} is conducted to get the final prediction at input resolution.

\subsection{Stacked Cascades for Inference \label{Infer}}
As is discussed in previous sections, during training we use a three-level feature pyramid at fixed resolutions to do hierarchical refinement. However, for images of higher resolution as input, more downsampling should be done in order to enlarge the receptive field for extracted features and correlation computation. But for small objects with large displacement in high resolution images, features in these regions may suffer from deterioration with direct downsampling.
To solve this problem, we designed a stacked cascaded architecture with shortcuts for inference. Specifically, we downsample the image pair in advance constructing an image pyramid and feed them into the same trained feature extraction network to take advantage of multi-level context. 
An overview of the stacked cascade architecture is shown on the right of Fig.~\ref{fig2:Architecture}, where skip connections in the same stage are not displayed for brevity. For a specific stage of the stacked cascades (denoted as rows in Fig.~\ref{fig2:Architecture}), all the RUMs in that stage will be used together with the last RUM in the stage of higher resolution. All stages of the stacked cascades share the same weight during training, so no fine-tuning is needed.

\subsection{Loss Function}
For each stage $s \in \{\frac{1}{16}, \frac{1}{8}, \frac{1}{4}\}$ of our
feature pyramid, we resize the sequence of output $\{\mathbf{f}_{i}^{s}, \cdots,
\mathbf{f}_{n}^{s}\}$ to the full prediction resolution with the upsampling operator
$\mu_{s}$, and use the exponentially weighted $l_1$ distance similar to RAFT
\cite{teed2020raft} as the loss function (with $\gamma$ set to 0.9). Given ground truth disparity
$\mathbf{d}_{\mathrm{gt}}$, the total loss is defined as:

\begin{equation}
\mathcal{L} = \sum_{s} \sum_{i=1}^{n} 
\gamma^{n - i} || \mathbf{d}_{\mathrm{gt}} - \mu_{s}(\mathbf{f}_{i}^{s}) ||_1
\end{equation}

\subsection{Synthetic Training Data}

\begin{figure}
    \centering
    \begin{subfigure}[b]{0.24\linewidth}
        \includegraphics[width=\linewidth]{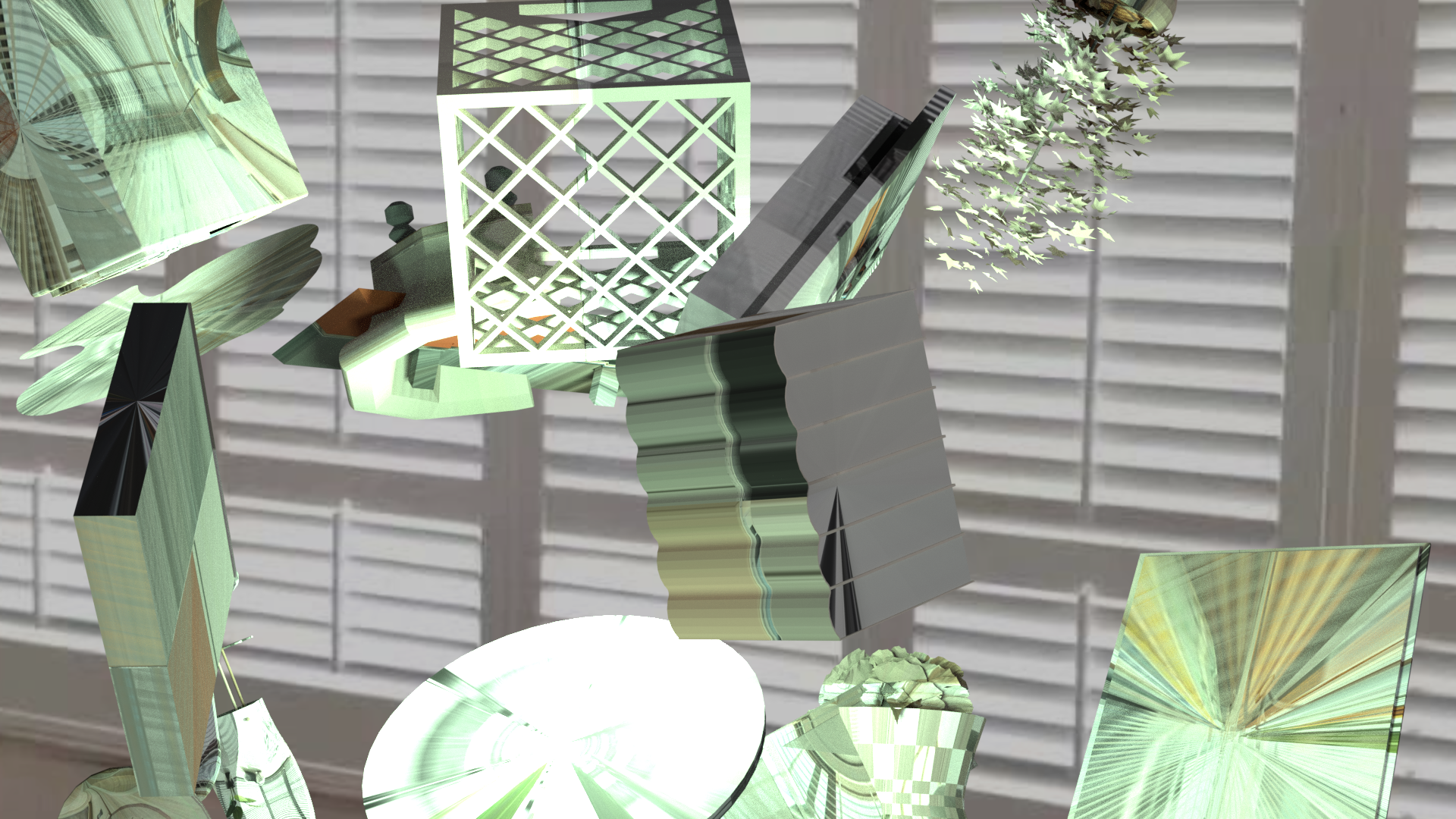}
    \end{subfigure}
    \begin{subfigure}[b]{0.24\linewidth}
        \includegraphics[width=\linewidth]{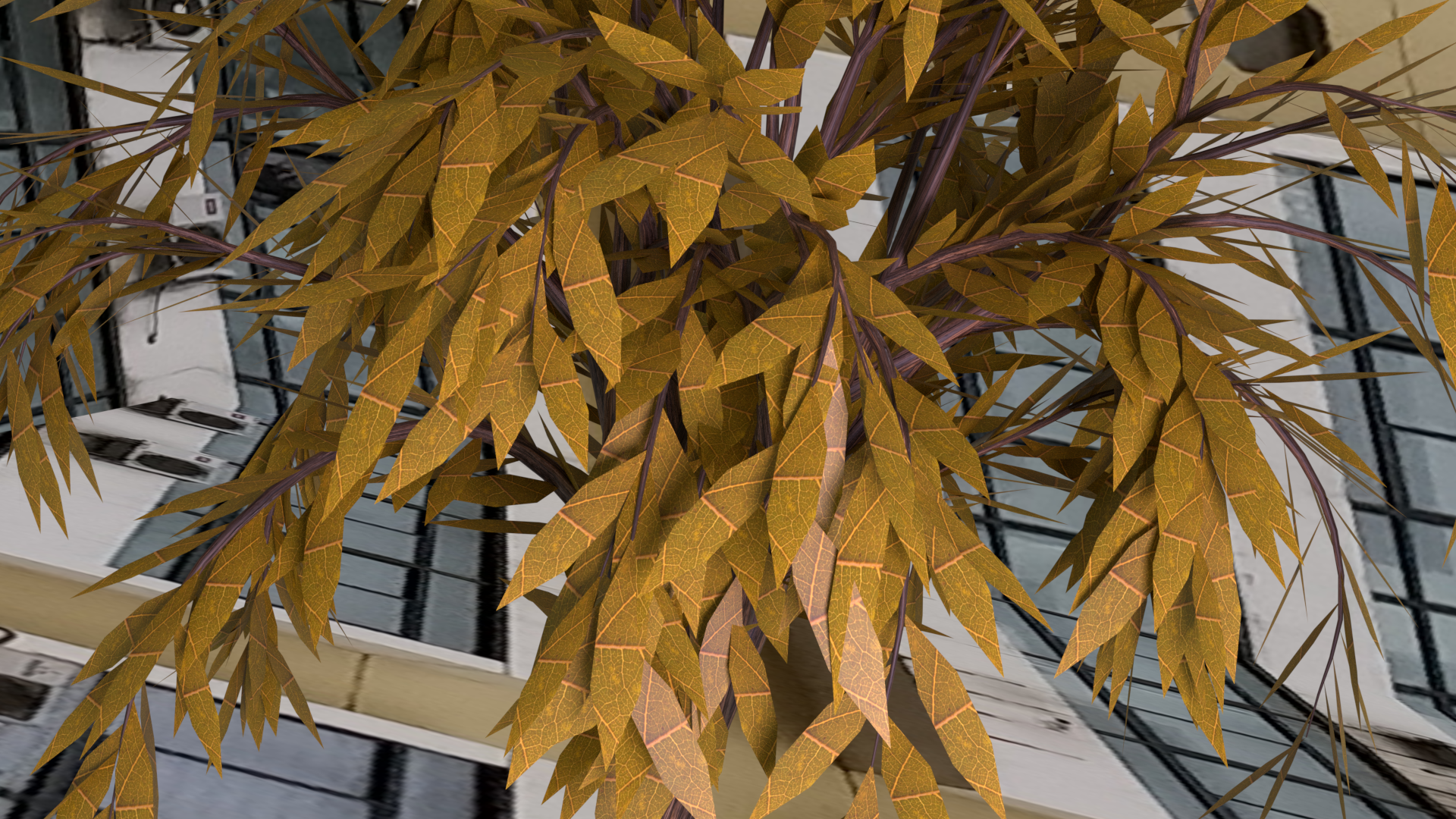}
    \end{subfigure}
    \begin{subfigure}[b]{0.24\linewidth}
        \includegraphics[width=\linewidth]{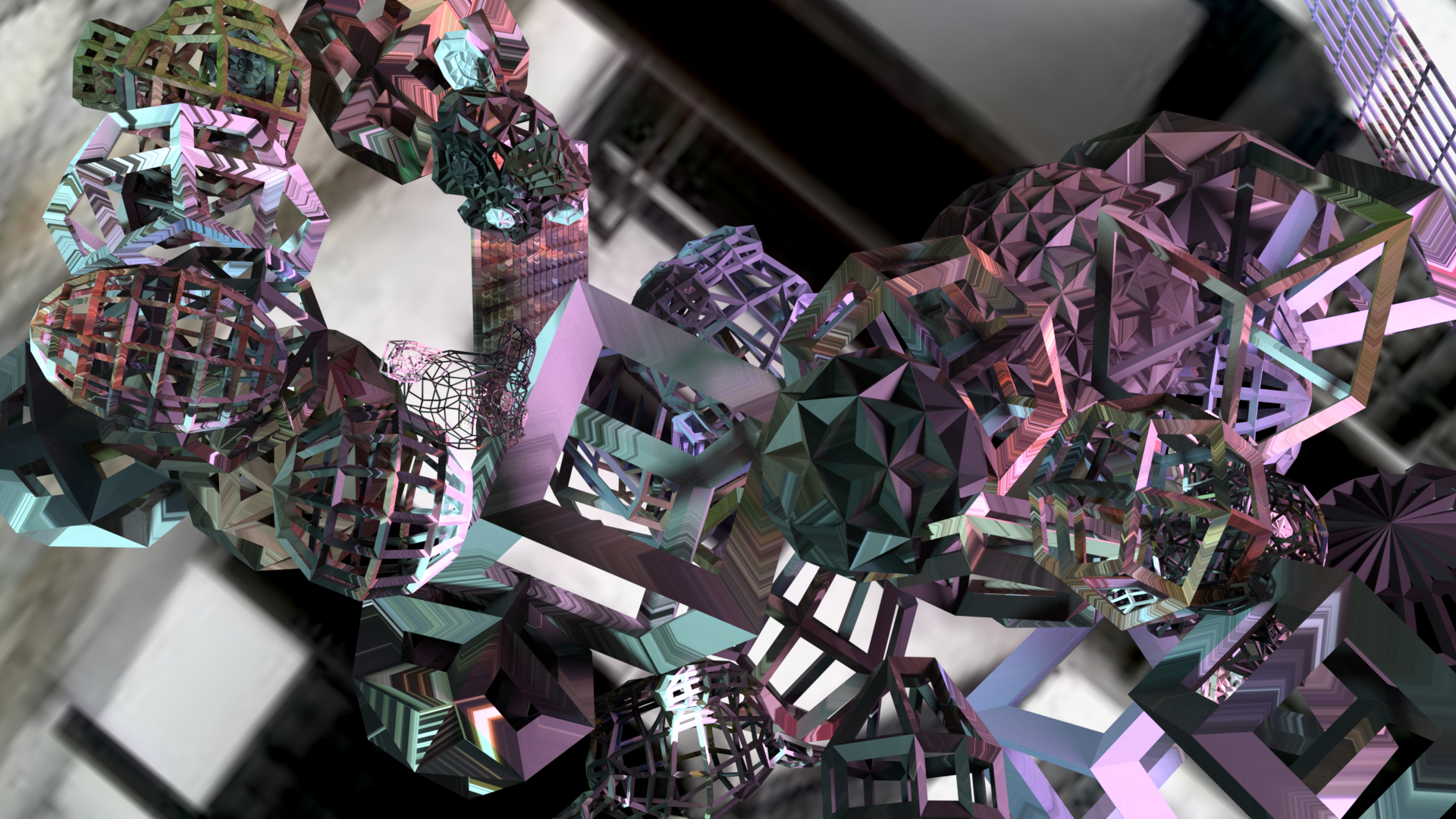}
    \end{subfigure}
    \begin{subfigure}[b]{0.24\linewidth}
        \includegraphics[width=\linewidth]{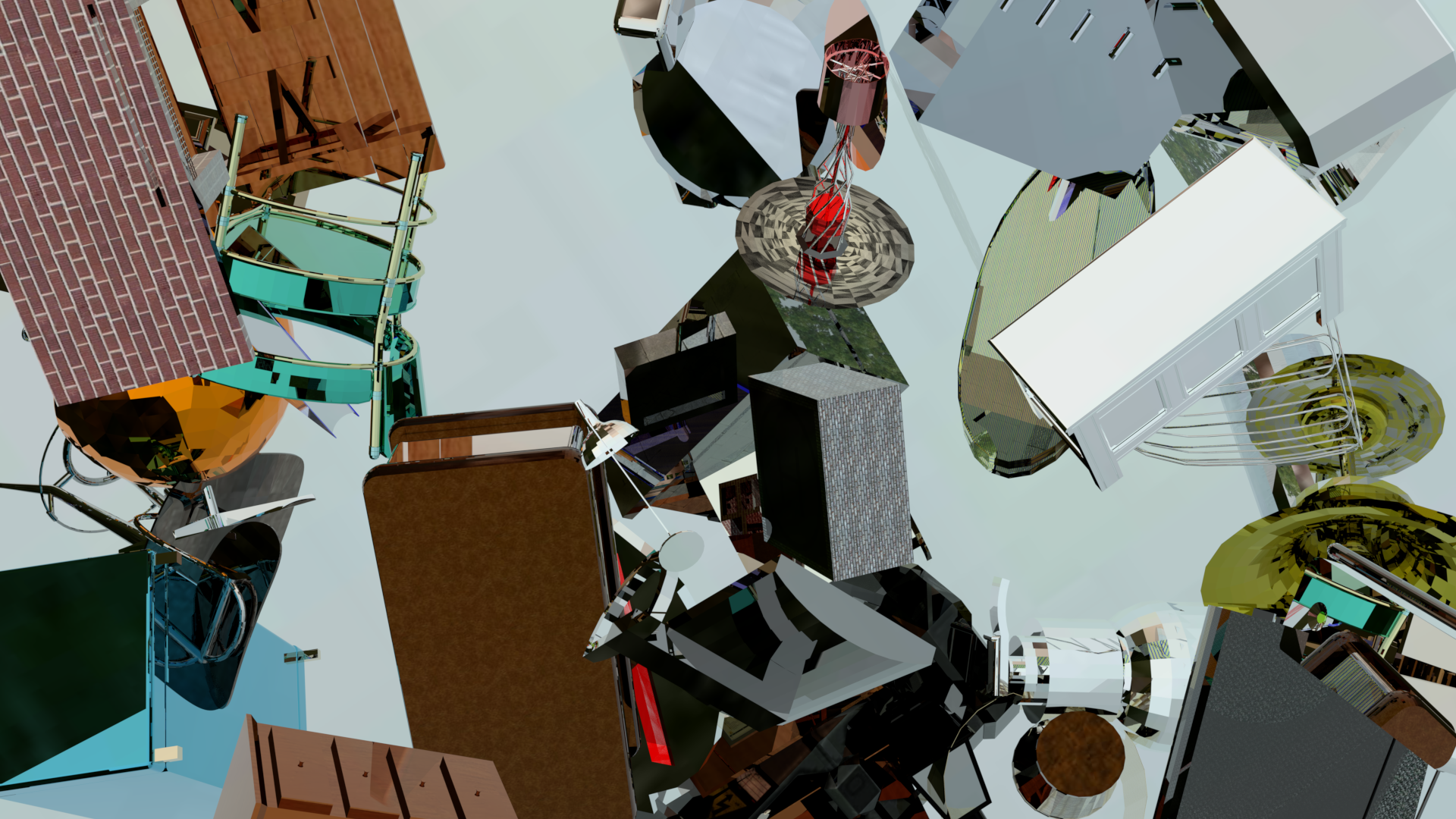}
    \end{subfigure}
    \\
    \begin{subfigure}[b]{0.24\linewidth}
        \includegraphics[width=\linewidth]{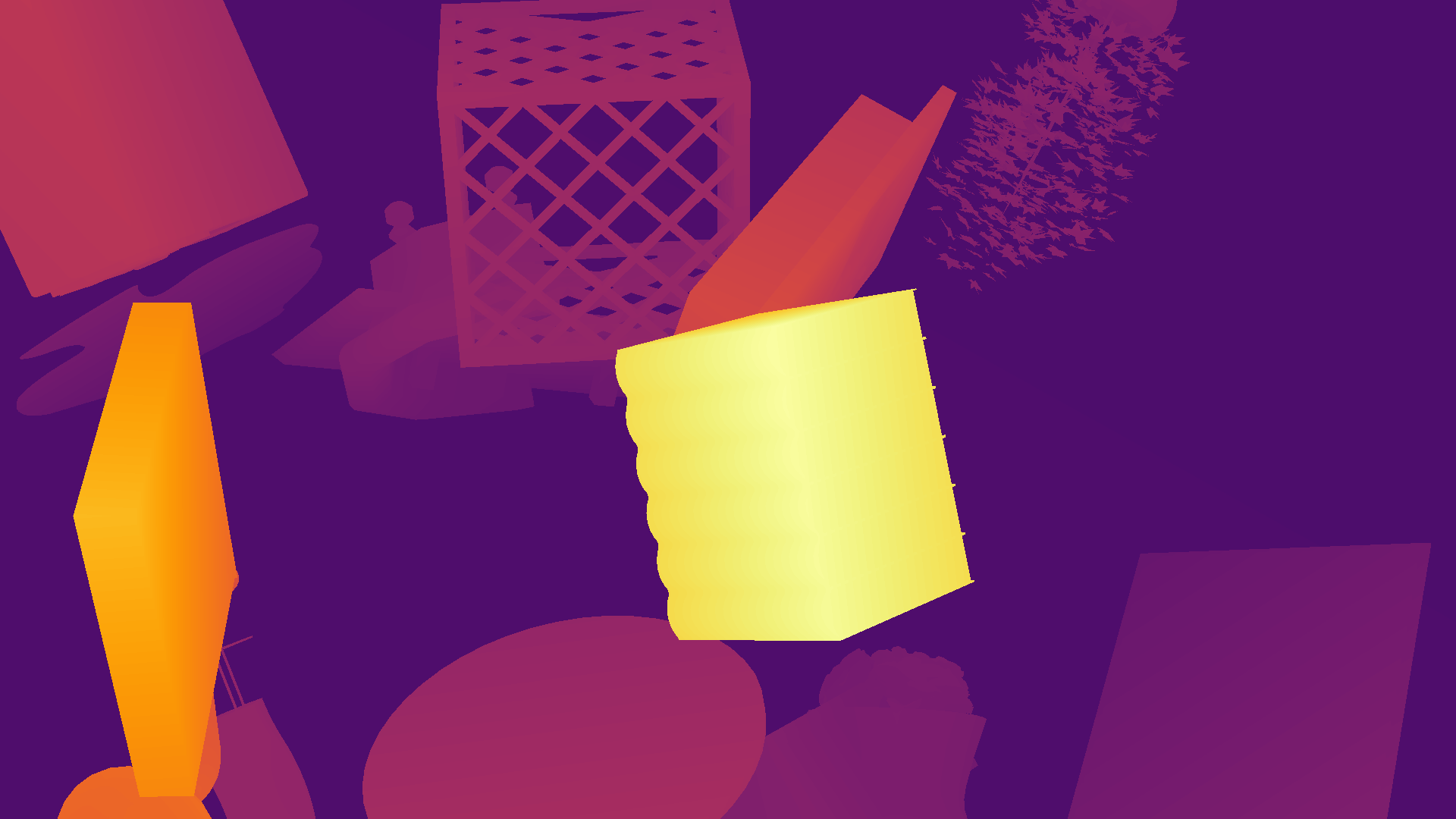}
        % \caption{Shapenet}
    \end{subfigure}
    \begin{subfigure}[b]{0.24\linewidth}
        \includegraphics[width=\linewidth]{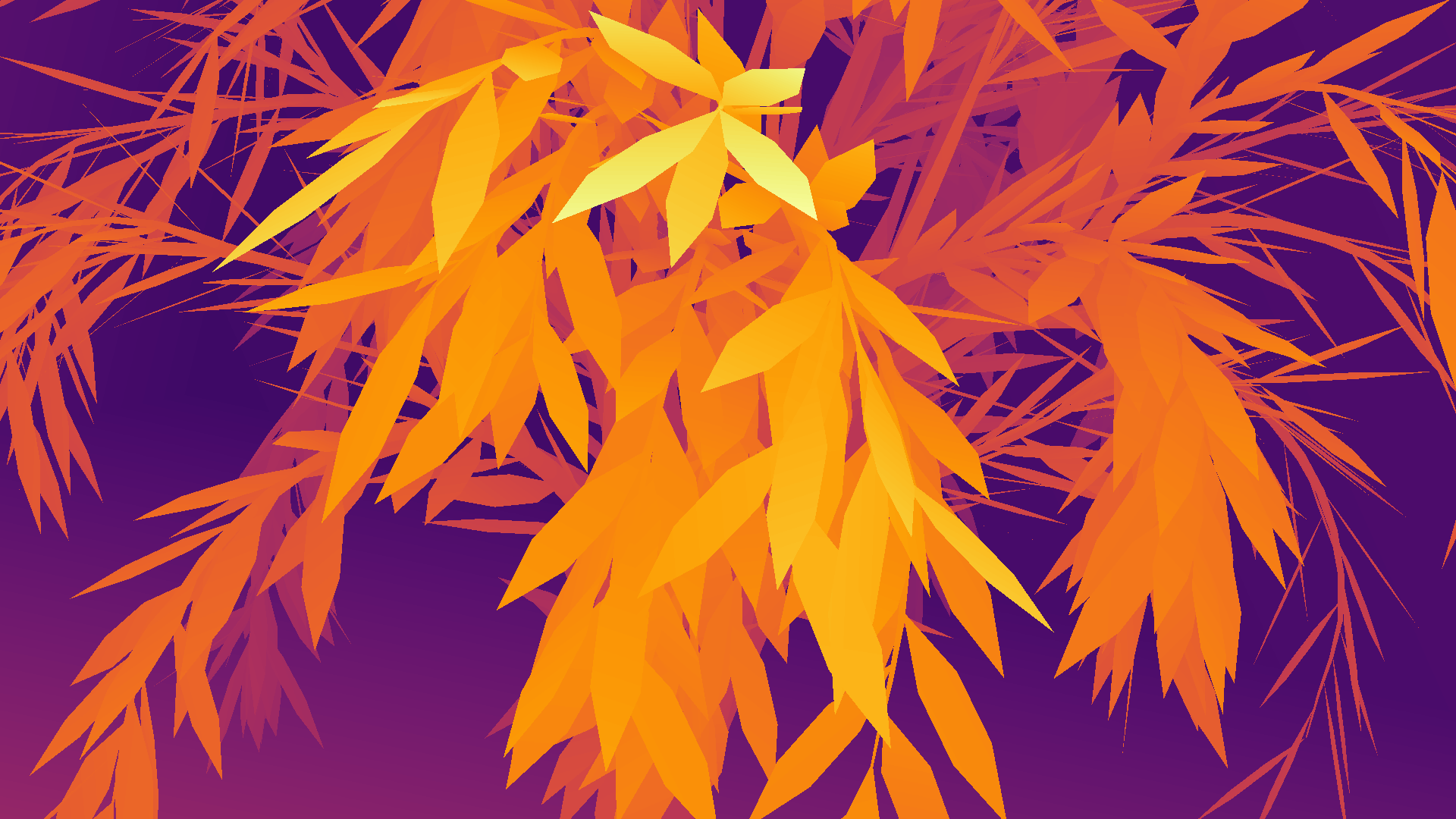}
        % \caption{Tree}
    \end{subfigure}
    \begin{subfigure}[b]{0.24\linewidth}
        \includegraphics[width=\linewidth]{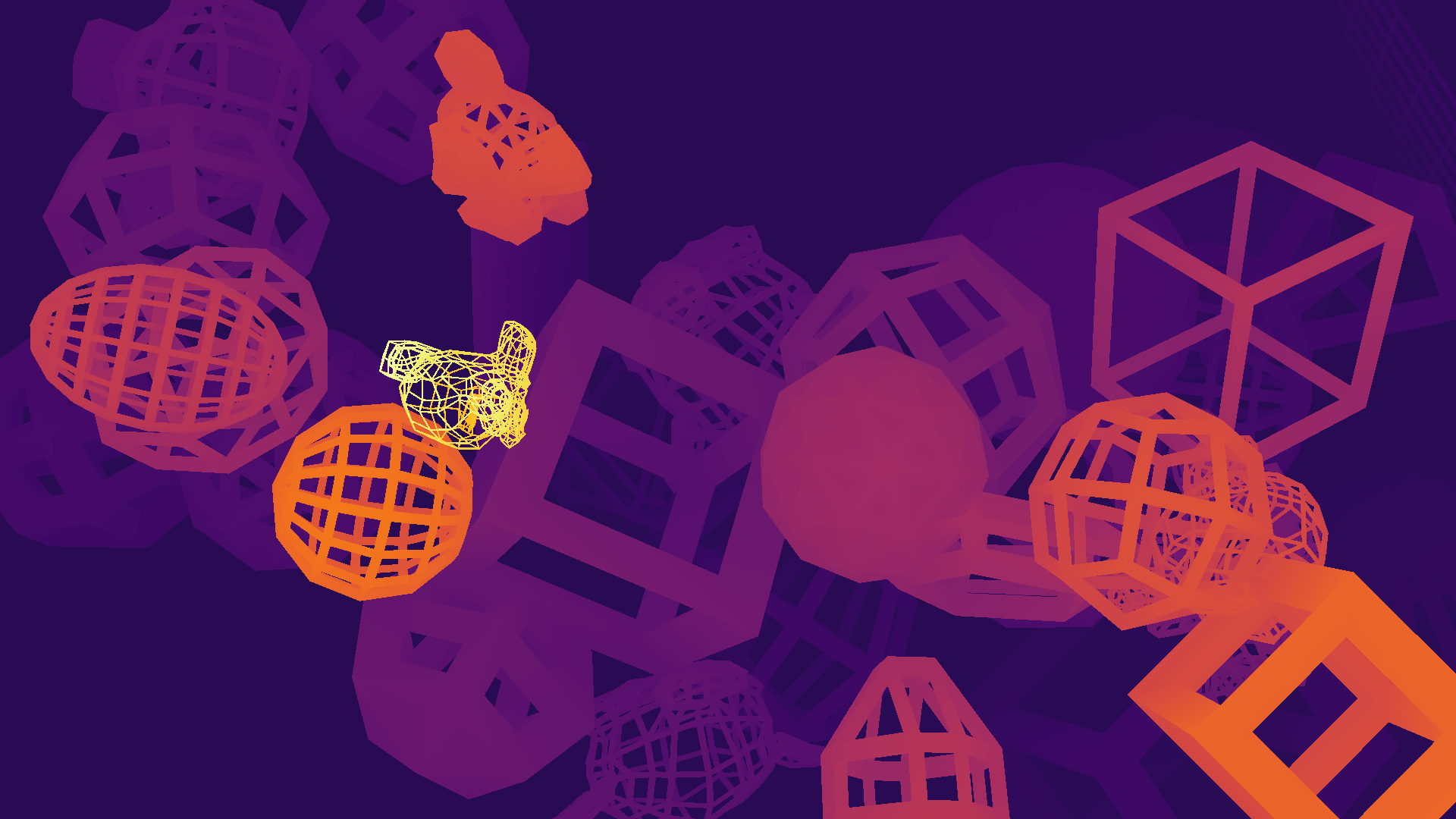}
        % \caption{Hole}
    \end{subfigure}
    \begin{subfigure}[b]{0.24\linewidth}
        \includegraphics[width=\linewidth]{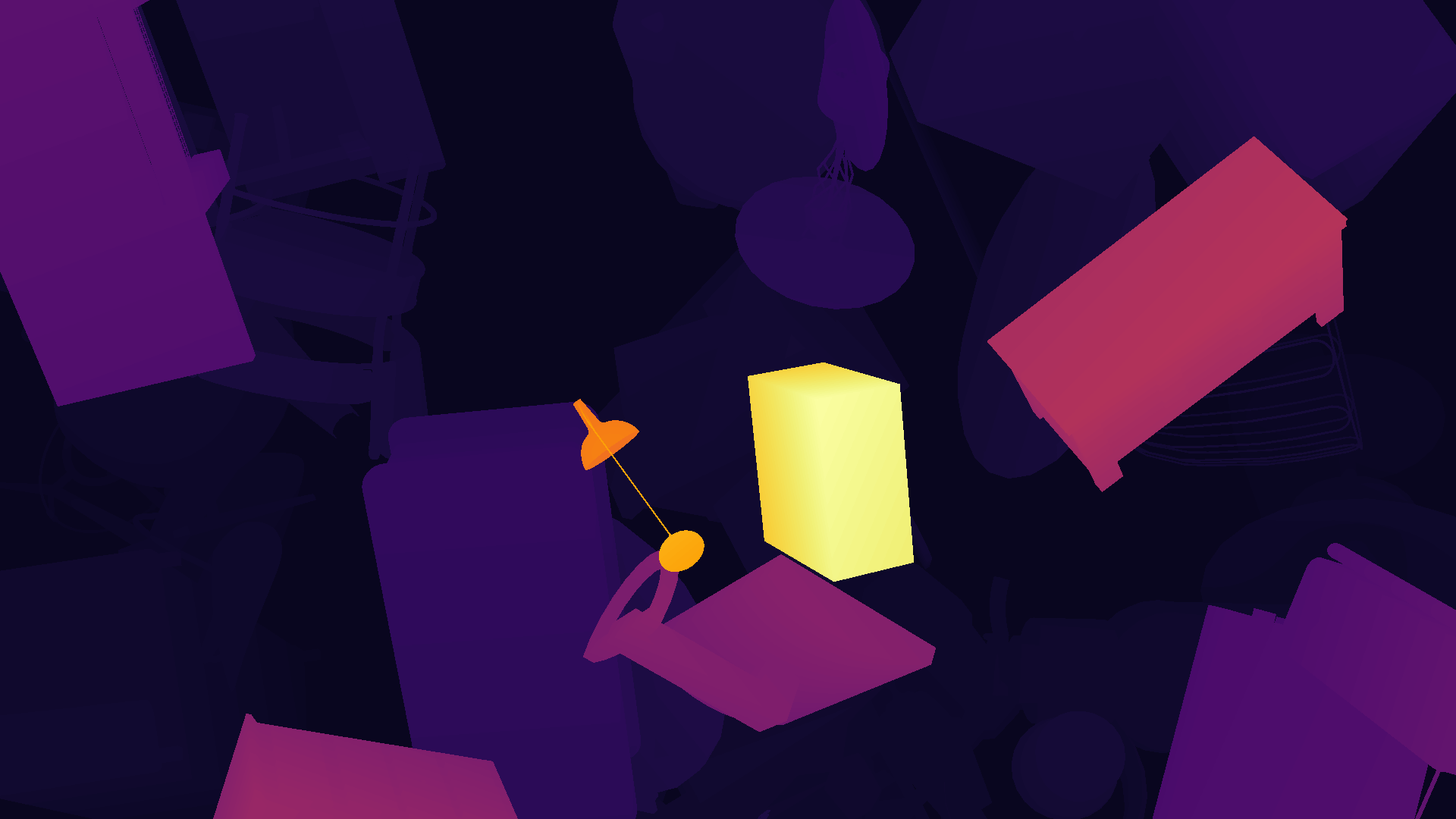}
        % \caption{Reflective}
    \end{subfigure}
    \\
    \vspace{-0.5em}
    \caption{Example image-disparity pairs of our synthetic data featuring various shapes and textures (repetitive-texture, reflective non-texture surface, etc.)}
    \label{fig:synthetic-data}
\end{figure}
% \vspace{-5mm}

Compared to previous synthetic datasets, our
data generating pipeline devotes extra attention to challenging cases in real-world scenes, and features various enhancements. We make use of Blender \cite{blender} to generate our synthetic training data. Each scene consists of left-right image pairs and the corresponding pixel-accurate dense disparity map, captured with dual virtual cameras and customarily positioned objects. Our major design considerations are described as below, with some examples shown in Fig.~\ref{fig:synthetic-data}.

\textbf{Shape.} We diversify the shapes of the models used as the main scene
content with multiple sources: 1) The ShapeNet \cite{shapenet}
dataset with over 40,000 3D models of common objects with varied shapes, forming
our basic source of content. 2) Blender's sapling tree gen add-on,
providing fine-detailed and cluttered disparity maps. 3) We use
blender's internal basic shapes combined with the wireframe modifier to
generate models for challenging scenes featuring holes and open-work
structures.

\textbf{Lighting and texture.} We place different types of lights with
random color and luminance at random position inside the scene, resulting in a
complex lighting environment. Real world images are used as textures for
objects and scene background, particularly hard scenes containing repeated patterns or lacking visible features. Additionally, we exploit the light tracing ability of Blender's Cycles renderer and randomly set objects as transparent or with metallic reflection, in order to cover real-world scenes with similar attributes.

\textbf{Disparity distribution.} To cover different baseline settings, we make
efforts to ensure the disparity of the generated data distributes smoothly
within a wide range. We put objects within a frustum-shaped space
formed by the cameras' field of view and a max
distance. The exact position of each object is randomly chosen from a probability distribution, then the object is scaled according to its distance to prevent blocking the view. This practice results in a randomized but controllable disparity distribution.

\section{Experiments}

\subsection{Datasets and Evaluation Metrics}
% We evaluate our network on multiple commonly used public stereo
% matching benchmarks, as well as real-world stereo photos captured with
% dual-camera smartphones.

% \subsubsection{Public Datasets}

We evaluate our method on three popular public benchmarks.
Middlebury 2014 \cite{scharstein2014high} provides 23 high-resolution image pairs
under different lighting environments. Captured with large-baseline stereo
cameras, the maximum disparity in Middlebury can exceed 600 pixels. ETH3D
\cite{eth3d} consists of 27 monochrome stereo image pairs with disparity
sampled by a laser scanner, covering both indoor and outdoor scenes. KITTI
2012/2015 \cite{menze2015object} consists of 200 wide-angle stereo
image pairs of street views, with lidar-sampled sparse disparity ground truth.

In addition to our rendered dataset, we collect major public datasets for training, including Sceneflow \cite{mayer2016large},
Sintel \cite{sintel} and Falling Things \cite{fallingthings}. Sceneflow
contains 39k training pairs of multiple synthetic scene setups. Falling things
contains a large amount of images from scenes of household object
models. Sintel provides 1.2k stereo pairs from various synthetic
sequences. The other data sources we utilize are InStereo2K
\cite{bao2020instereo2k}, Carla \cite{deschaud2021kitti} and AirSim
\cite{airsim2017fsr}.

For evaluation, we follow the popular metrics including AvgErr  (average error), Bad2.0 
(percentage of pixels with disparity error larger than 2 pixels) \cite{middlebury, eth3d},
D1-all (percentage of disparity outlier pixels in left image) \cite{kitti}, etc.

\subsection{Implementation Details}
\textbf{Training. } Our network is implemented with Pytorch \cite{pytorch} framework. The model is trained on 8 NVIDIA GTX 2080Ti GPUs, with
a batch size of 16. The whole training process is set to 300,000
iterations. We use the Adam \cite{adam} optimizer with a standard learning
rate of 0.0004. We perform a warm-up process of 6,000 iterations at the
beginning of the training where the learning rate is linearly increased from 5\% to
100\% of the standard value. After 180,000 iterations, the learning rate
is linearly decreased down  to 5\% of the standard value towards the end of
the training process. The model is trained with an input size of $384 \times 512$.
All training samples undergo a set of augmentation operations before getting fed
into the model.

\textbf{Augmentation.}  To imitate the camera module inconsistencies and non-ideal rectification, we employ multiple data augmentation techniques for training. Firstly, we apply asymmetric chromatic augmentations for the two inputs respectively, including shifts in
brightness, contrast and gamma. To further enhance the
robustness for rectification error in real-world images, we conduct spatial augmentation applied
only to the right image: slightly random homography transformation and vertical shift at a very small
range ($<$ 2 pixels). 
To avoid mismatching in ill-posed regions, we use random rectangle occlusion patches with height and width between 50 and 100 pixels.
Finally, to fit input data from various sources into
the network's training input size, the group of stereo images and disparity
undergoes random resize and crop operations.

\subsection{Ablation Study}

\begin{table}[t]
  \centering
  \small
  \setlength{\tabcolsep}{3.pt}
  \begin{tabular}{lcccc}
    \toprule
    \multirow{2}{*}[-2pt]{Method} & \multicolumn{2}{c}{Middlebury} & \multicolumn{2}{c}{ETH3D} \\ 
    \addlinespace[-12pt] \\
    \cmidrule(lr){2-3} \cmidrule(lr){4-5}
    \addlinespace[-12pt] \\ 
    & Bad 2.0 & AvgErr & Bad 1.0 & Avgerr \\
    \midrule
    
    2D all-pairs \cite{teed2020raft} & 47.38 & 5.62 & 6.17 & 0.38 \\
    1D all-pairs \cite{lipson2021raft} & 44.41 & 4.93 & 6.03 & 0.38 \\
    1D local & 19.87 & 3.03 & 3.13 & 0.28 \\
    2D local & 20.70 & 2.99 & 3.33 & 0.29 \\
    1D+2D local & 19.23 & 3.01 & 3.05 & 0.28 \\
    1D local, 2 levels & 13.84 & 2.24 & 2.35 & 0.23 \\
    2D local, 2 levels & 14.07 & 2.15 & \underline{2.09} & 0.23 \\
    1D+2D local, 2 levels & {\bf 12.48} & \underline{1.99} & 2.20 & \underline{0.22} \\
    1D+2D local, 3 levels & \underline{12.67} & {\bf 1.80} & {\bf 2.01} & {\bf 0.21} \\
    \midrule
    \midrule
    w/o def. \& group. \& atten. & 6.86 & 1.11 & 1.26 & 0.19 \\
    w/o deformable search  & 6.84 & 1.08 & 1.22 & 0.19 \\
    w/o group-wise correlation & 6.82 & 1.08 & \underline{1.20} & \underline{0.18} \\
    w/o attention & \underline{6.49} & \underline{1.07} & 1.22 & \underline{0.18} \\
    full method & {\bf 6.46} & {\bf 1.05} & {\bf 1.03} & {\bf 0.17} \\
    \bottomrule
  \end{tabular}
  \vspace{-5pt}
  \caption{
  Ablation study for RUMs. The top half is comparisons for different forms of correlation layers and different levels of cascades, trained on public datasets except Middlebury and ETH3D. And the bottom half is evaluation for key components in AGCL, trained on full datasets. }
  \label{tab:ablation}
%   \vspace{-10pt}
  
\end{table}

\begin{table}[t]
  \centering
  \small
  \setlength{\tabcolsep}{3.pt}
  \begin{tabular}{lccc}
    \toprule
    \multirow{2}{*}[-2pt]{Method} &
    \multirow{2}{*}[-2pt]{Input size} &
    %% Middlebury \\ bad2.0 & Middlebury \\ avgerr & Smartphone \\ maxIoU \\
    \multicolumn{2}{c}{Middlebury}  \\ 
    \cmidrule(lr){3-4}
    & & Bad 2.0 & AvgErr \\ 
    \midrule
    single cascade & $768 \times 1024$ & 6.46 & 1.05 \\
    single cascade & $1024 \times 1536$ & 6.00 & 1.61  \\
    2 stacked cascades & $1024 \times 1536$ & 5.30 & 0.94 \\
    2 stacked cascades & $1536 \times 2048$ & {\bf4.53} & \underline{0.93} \\
    3 stacked cascades & $1536 \times 2048$ &  \underline{4.58} & {\bf 0.92} \\
    \bottomrule
  \end{tabular}
  \vspace{-5pt}
  \caption{Abalation study for stacked cascaded architecture during inference.}
  \label{tab:cascade}
\end{table}

In this section we evaluate our model on different settings to prove the effectiveness of the network components. Besides the ablation study for the stacked cascades, all evaluation resolutions are $768 \times 1024$.

\textbf{Correlation types.}
In order to compare the effect of different types of correlations, we replace our correlation layers with other forms. As shown in Tab.~\ref{tab:ablation}, both 2D and 1D all-pairs correlation used in \cite{teed2020raft} and \cite{lipson2021raft} lead to a substantial drop in accuracy compared with their local forms. When we replace the alternate local correlation with a single 2D or 1D correlation, it harms the final precision, which is more evident when the network contains more than 1 level of cascades because the rectification error increases with the resolution.

\textbf{Components in AGCL.}
As shown in the bottom half of Tab.~\ref{tab:ablation}, using a fixed correlation window without learned offsets degrades the accuracy which demonstrates the effectiveness of the adaptive mechanism. Replacing group correlation with single form and removing the local feature attention modules both deteriorate the accuracy. 

\textbf{Cascaded RUMs.}
We compare the performance of different numbers of cascade stages. As is shown in Tab.~\ref{tab:ablation}, using a single RUM without cascades leads to a substantial drop in precision. When changing the number of cascades, the prediction error decreases evidently when more levels of cascades are used while the correlation type keeps the same. This demonstrates the importance of our cascaded architecture.

\textbf{Stacked cascades.}
During inference, we feed the cascades using different levels of image pyramid as input while sharing the same trained parameters. 
We compare the performance of different stages of cascades with various of resolutions on Middlebury. As shown in Tab.~\ref{tab:cascade}, the prediction error increases with the input size when only a single cascade is used. Multi-level input helps to reduce the error substantially, which demonstrates that our stacked cascades scheme enjoys a great improvement for disparity accuracy.

\begin{figure}
    \centering
    \includegraphics[width=\linewidth]{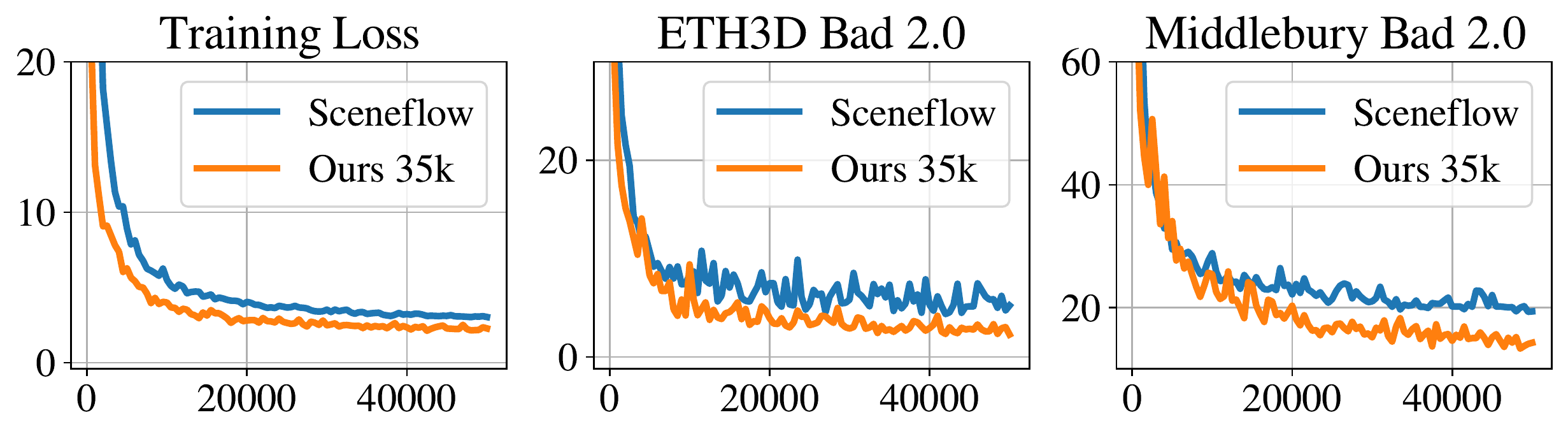}
    \\
    \vspace{-0.5em}
    \caption{Training loss and ETH3D / Middlebury validation error of models trained with Sceneflow and our synthetic dataset.}
    \label{fig:dataset_compare}
\end{figure}

\textbf{New synthetic data.}
To analyze the effectiveness of our proposed synthetic data, we sample 35,000
pairs of images from our training dataset and compare with similar-sized
Sceneflow \cite{mayer2016large}. Both datasets are used to train our model with the same augmentation for
50,000 iterations. As shown in Fig.~\ref{fig:dataset_compare}, our synthetic
data results in lower training loss and better performance in both ETH3D and
Middlebury validation data. This demonstrate that our dataset is more advantageous in domain generalization.

\begin{figure}
    \centering
    \begin{subfigure}[b]{0.24\linewidth}
        \includegraphics[width=\linewidth]{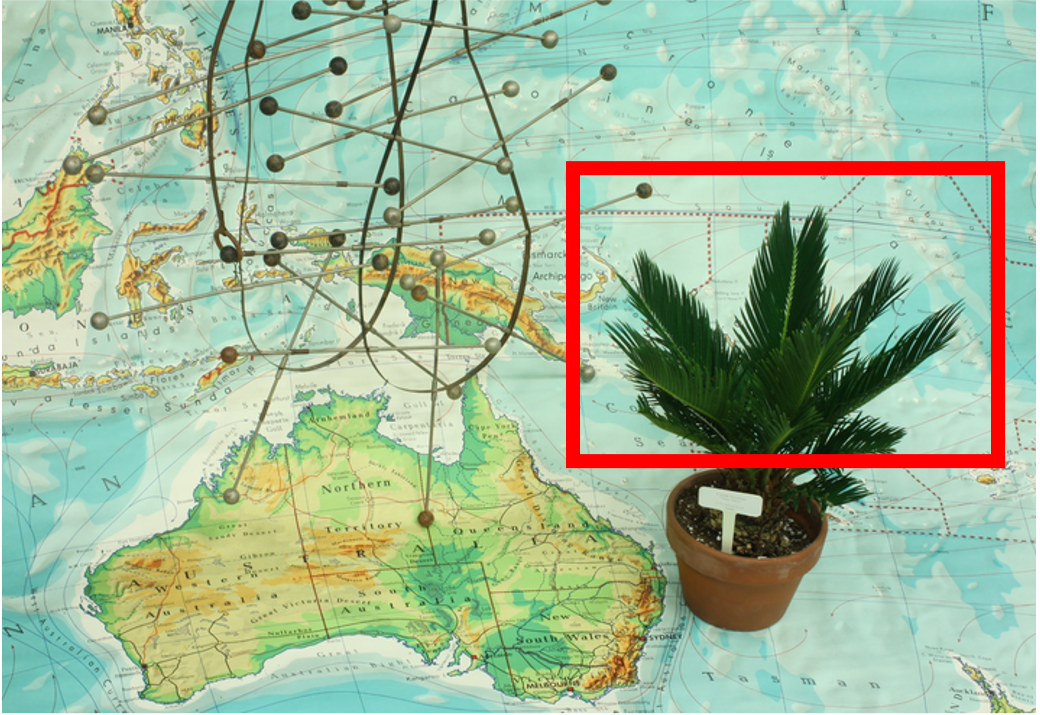}
    \end{subfigure}
    \begin{subfigure}[b]{0.24\linewidth}
        \includegraphics[width=\linewidth]{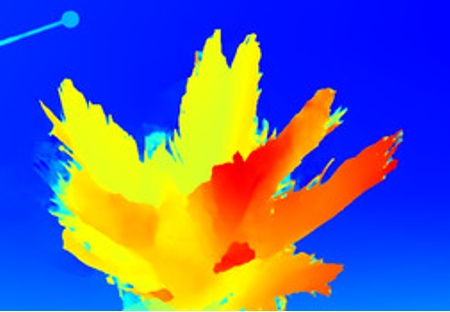}
    \end{subfigure}
    \begin{subfigure}[b]{0.24\linewidth}
        \includegraphics[width=\linewidth]{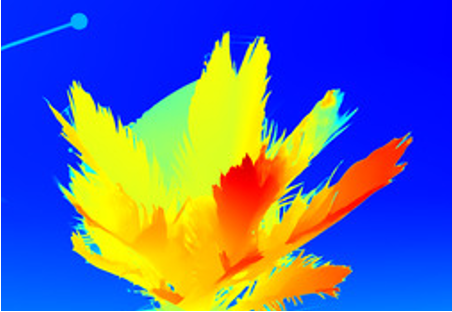}
    \end{subfigure}
    \begin{subfigure}[b]{0.24\linewidth}
        \includegraphics[width=\linewidth]{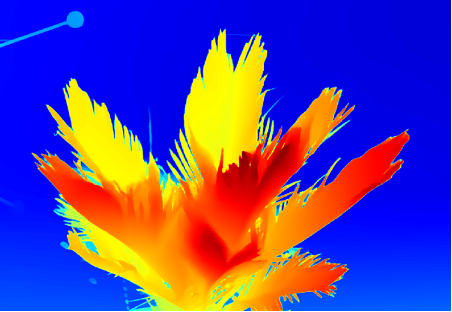}
    \end{subfigure}
    \\
    \vspace{0.15em}
    \begin{subfigure}[b]{0.24\linewidth}
      \includegraphics[width=\linewidth]{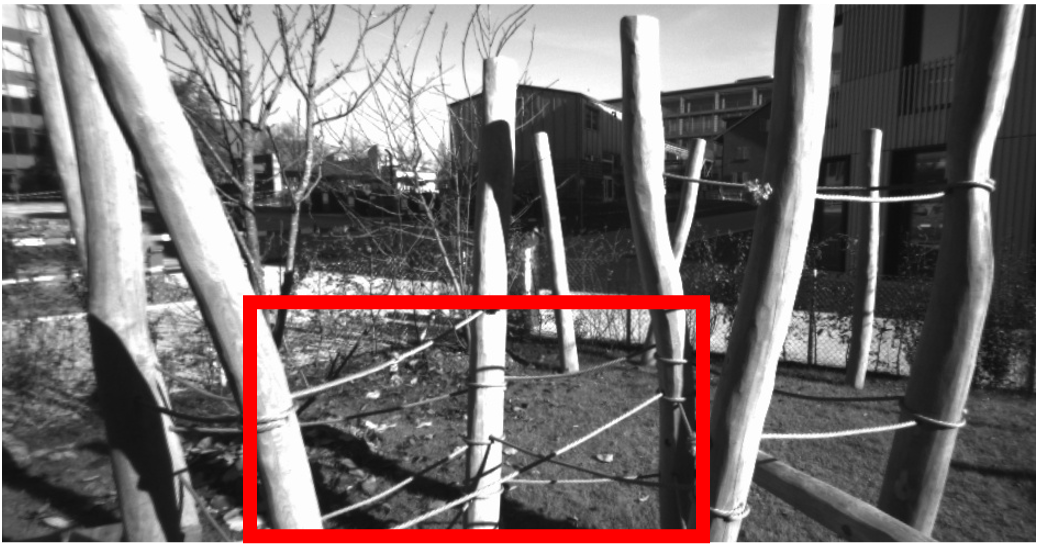}
      \caption{Left image}
    \end{subfigure}
    \begin{subfigure}[b]{0.24\linewidth}
      \includegraphics[width=\linewidth]{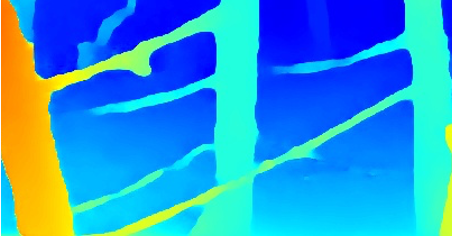}
      \caption{HITNet}
    \end{subfigure}
    \begin{subfigure}[b]{0.24\linewidth}
      \includegraphics[width=\linewidth]{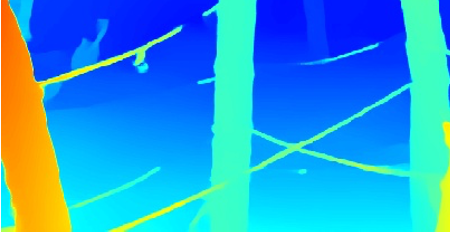}
      \caption{ RAFT-Stereo}
    \end{subfigure}
    \begin{subfigure}[b]{0.24\linewidth}
      \includegraphics[width=\linewidth]{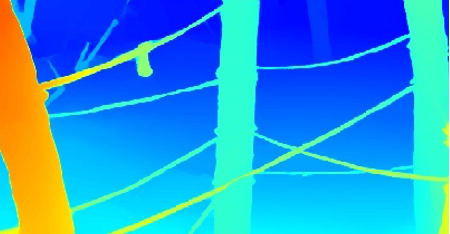}
      \caption{Ours}
    \end{subfigure}
    \\
    \vspace{-0.5em}
    \caption{Visual comparisons on Middelbury and ETH3D with HITNet \cite{tankovich2021hitnet} and RAFT-Stereo \cite{lipson2021raft}.}
    \label{fig:compare_mb_eth}
\end{figure}

\begin{figure}
    \centering
    \begin{subfigure}[b]{0.48\linewidth}
        \includegraphics[width=\linewidth]{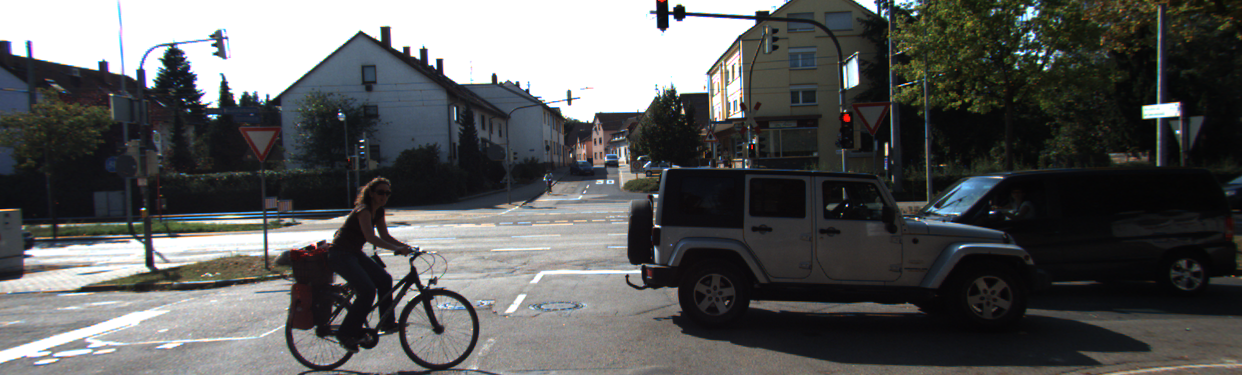}
        \caption{Left image}
        \label{fig:compare_kitti15_left}
    \end{subfigure}
    \begin{subfigure}[b]{0.48\linewidth}
        \includegraphics[width=\linewidth]{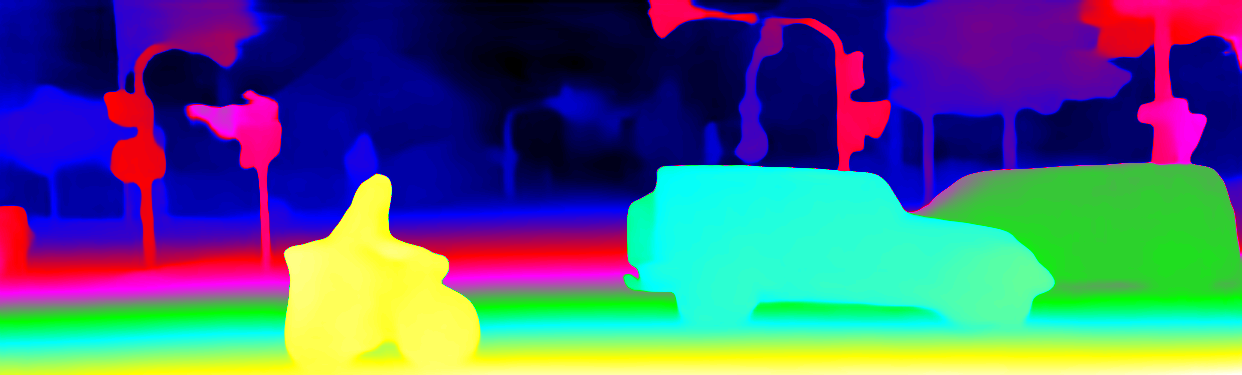}
        \caption{LEAStereo\cite{cheng2020hierarchical}}
        \label{fig:compare_kitti15_LEAStereo}
    \end{subfigure}
    \\
    \begin{subfigure}[b]{0.48\linewidth}
        \includegraphics[width=\linewidth]{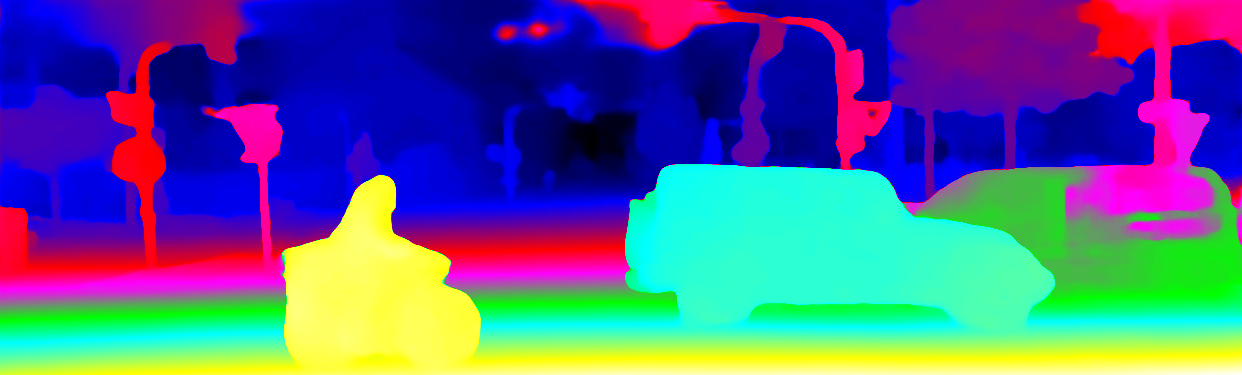}
        \caption{AANet\cite{xu2020aanet}}
        \label{fig:compare_kitti15_AANet}
    \end{subfigure}
    \begin{subfigure}[b]{0.48\linewidth}
        \includegraphics[width=\linewidth]{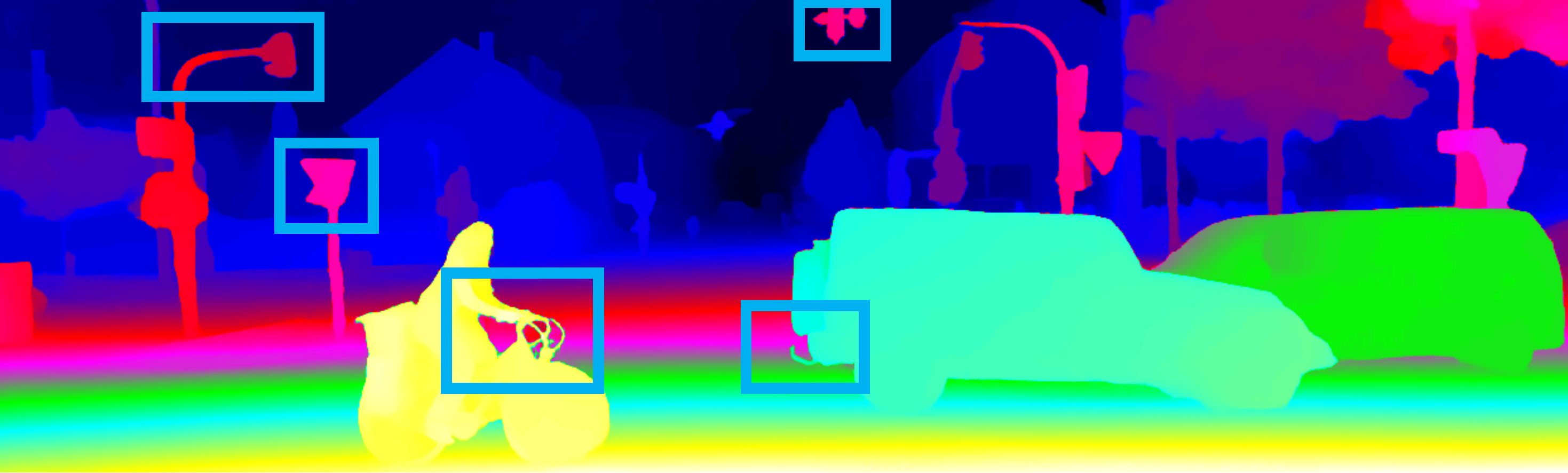}
        \caption{Ours}
        \label{fig:compare_kitti15_cascade}
    \end{subfigure}
    \vspace{-0.5em}
    \caption{Visual comparisons with other methods on one case of KITTI 2015 test set. Our method preserves more details.}
    \label{fig:compare_kitti15}
\end{figure}

\begin{figure*}[t]
  \begin{center}%
    \newcommand{\galleryRowCompare}[7]{
      \vspace*{-0.075cm}
      \includegraphics[width=0.2\textwidth]{compare_hardcase/#1} &
      \includegraphics[width=0.2\textwidth]{compare_hardcase/#2} &
      \includegraphics[width=0.2\textwidth]{compare_hardcase/#3} &
      \includegraphics[width=0.2\textwidth]{compare_hardcase/#4} &
      \includegraphics[width=0.2\textwidth]{compare_hardcase/#5} &
      \includegraphics[width=0.2\textwidth]{compare_hardcase/#6} &
      \includegraphics[width=0.2\textwidth]{compare_hardcase/#7}
      \tabularnewline 
    }
  \resizebox{\linewidth}{!}{%
    \setlength{\tabcolsep}{0.7pt}%
    \begin{tabular}{ccccccc}%
      Image &
      AANet \cite{xu2020aanet} &
      HSMNet \cite{wang2020improving} &
      GwcNet \cite{guo2019group} &
      LEAStereo \cite{cheng2020hierarchical} &
      RAFT-Stereo \cite{lipson2021raft} &
      Ours \\
    \galleryRowCompare{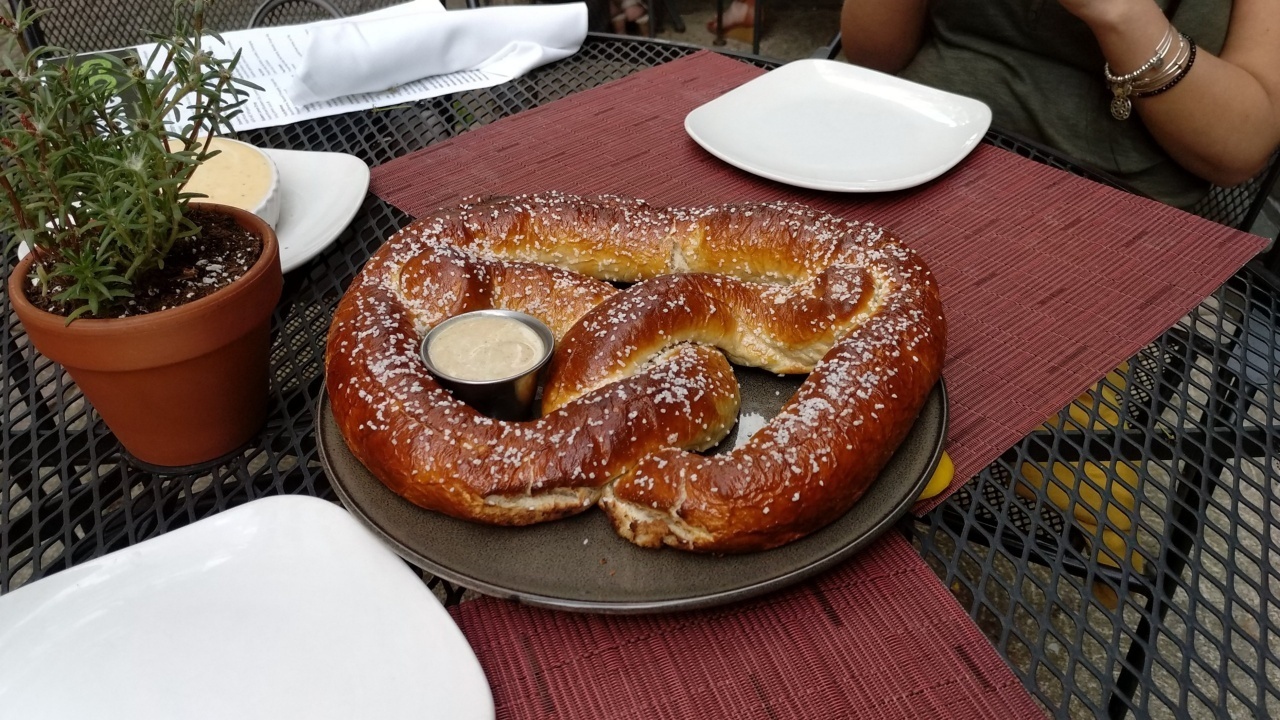}{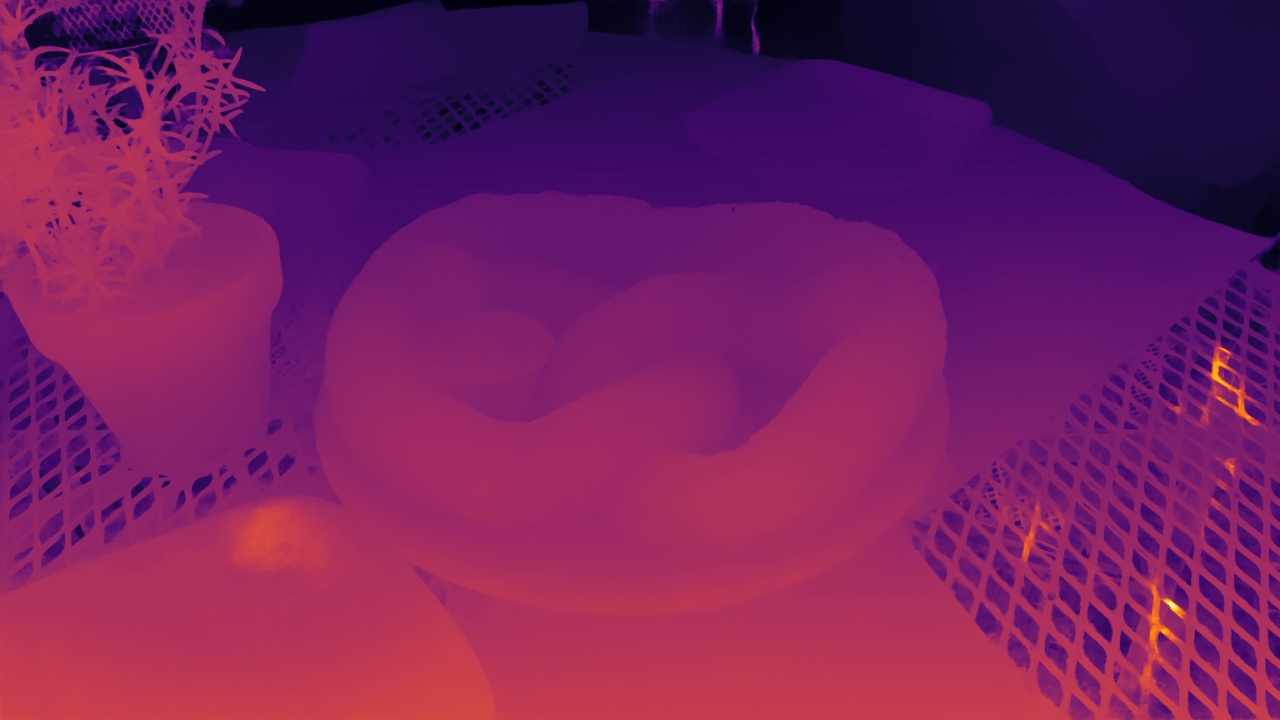}{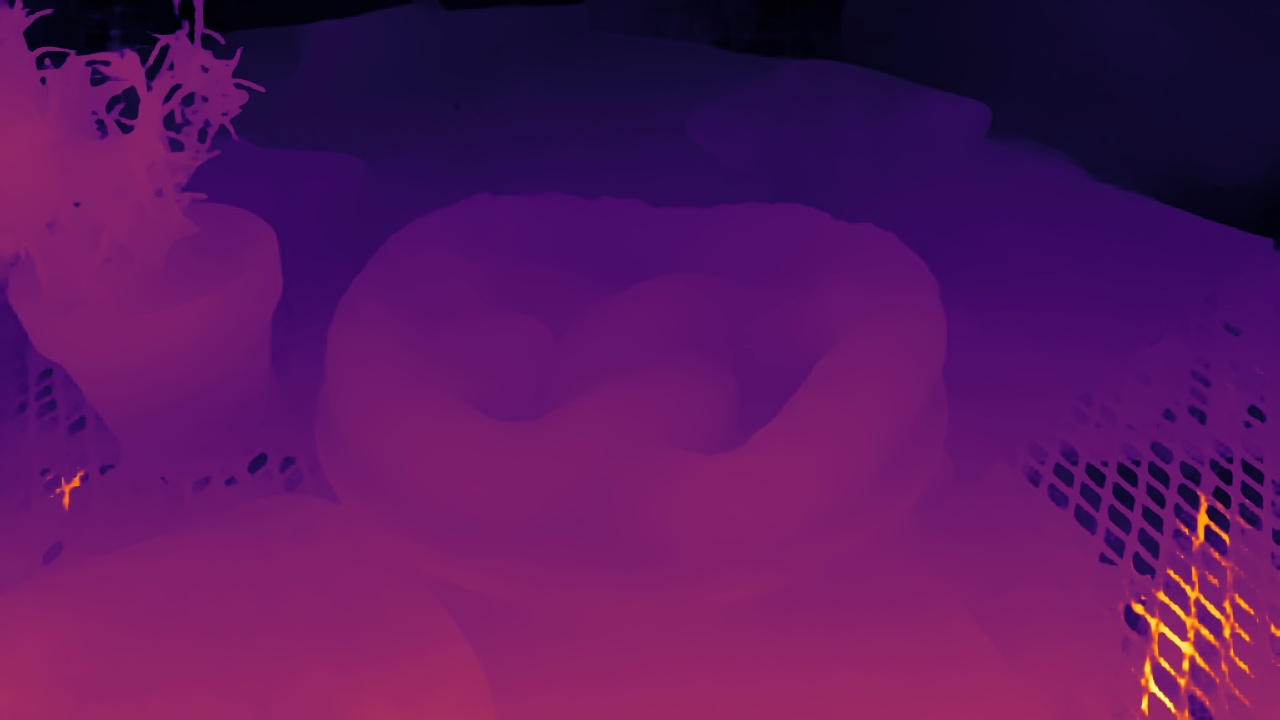}{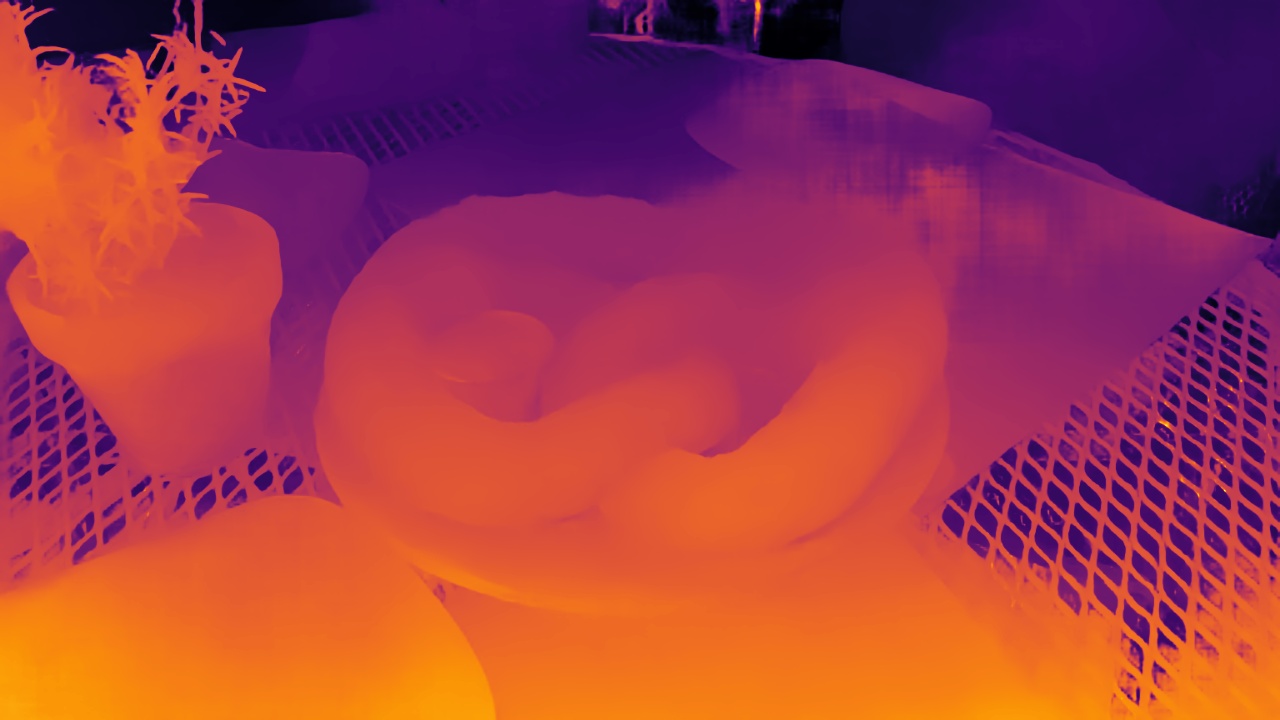}{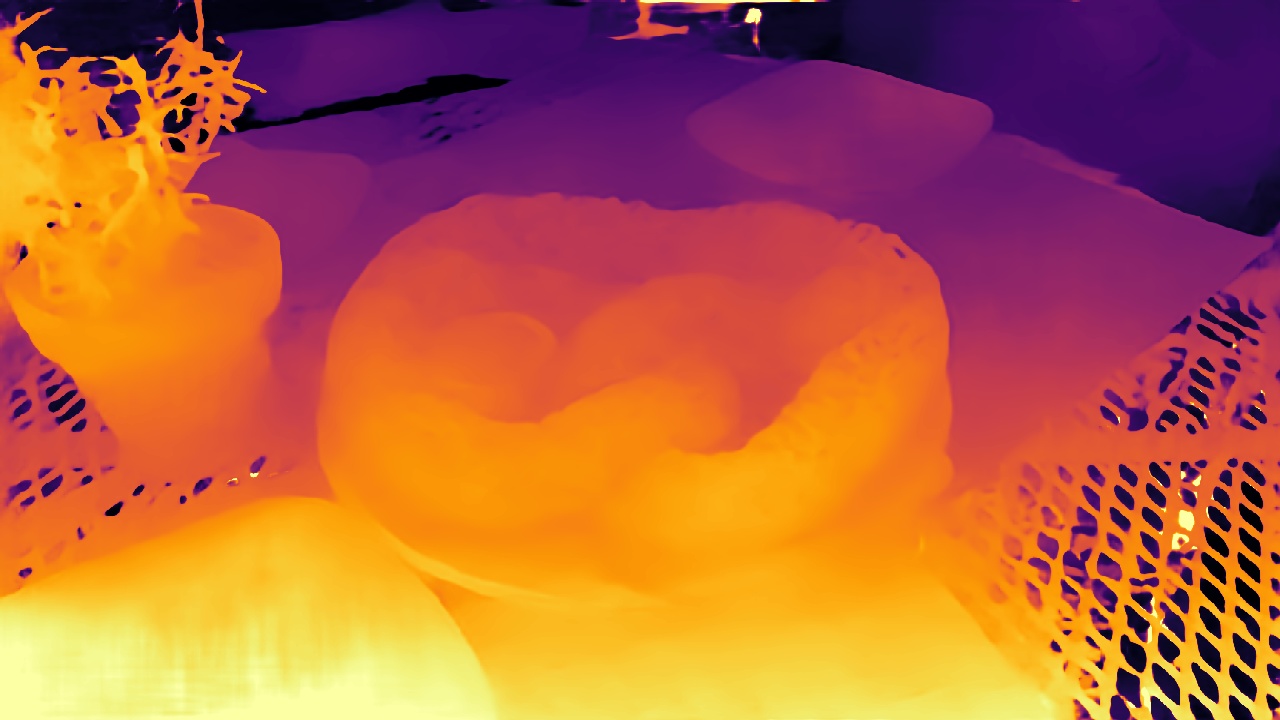}{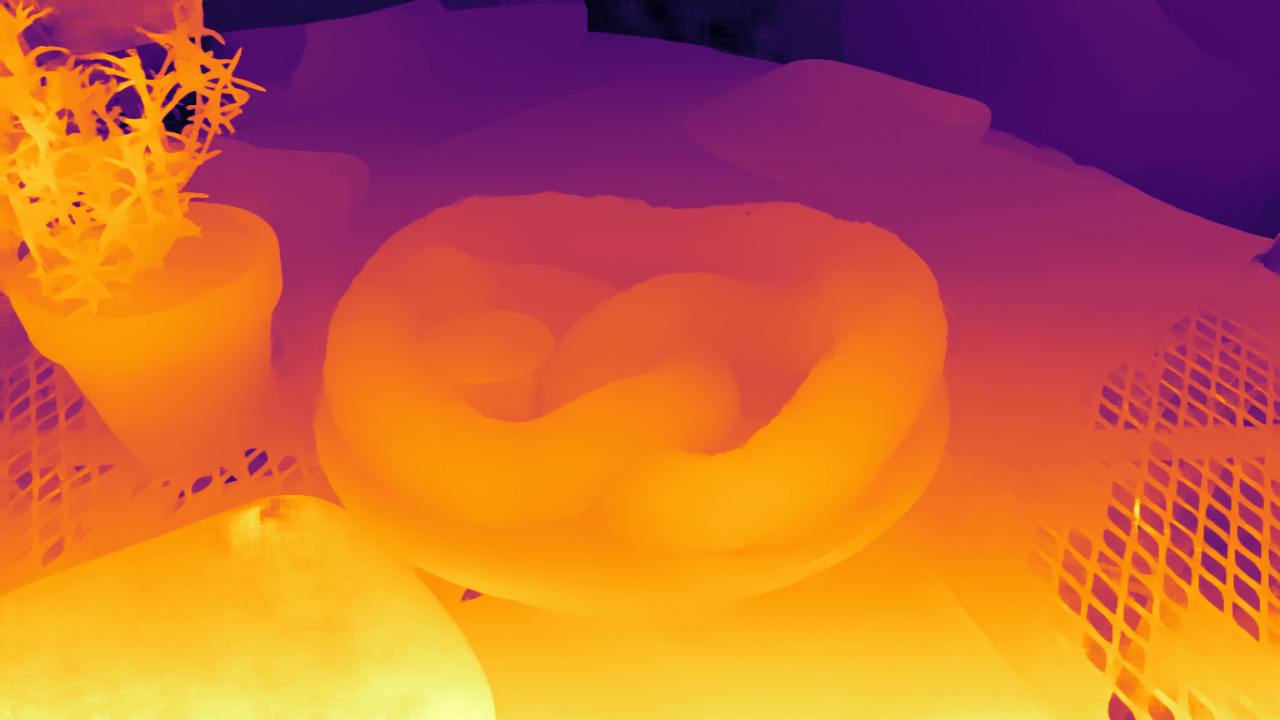}{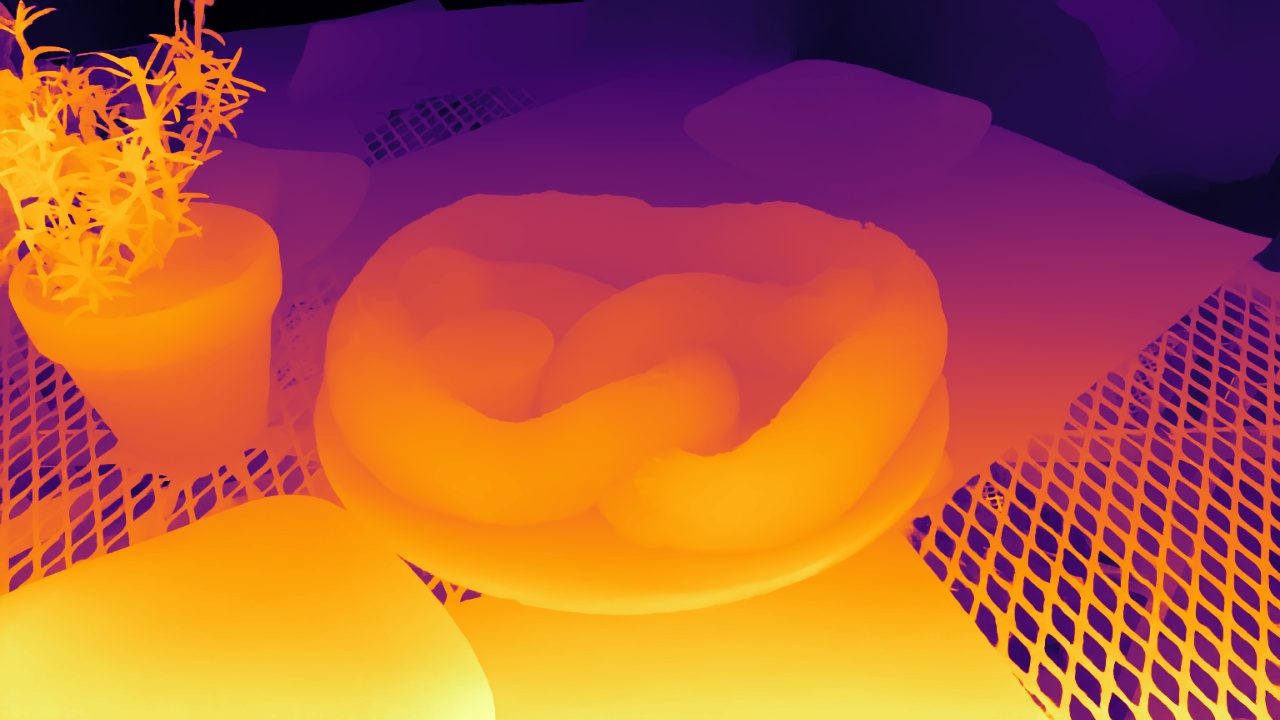}%
    \galleryRowCompare{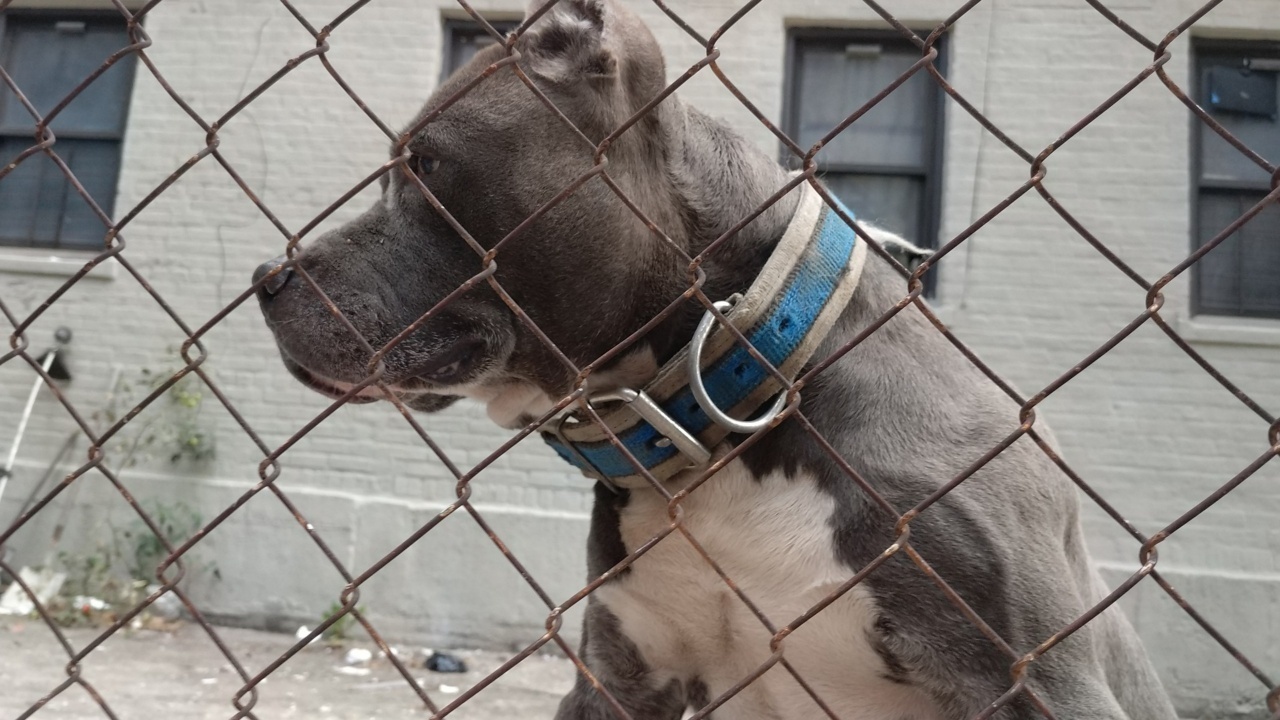}{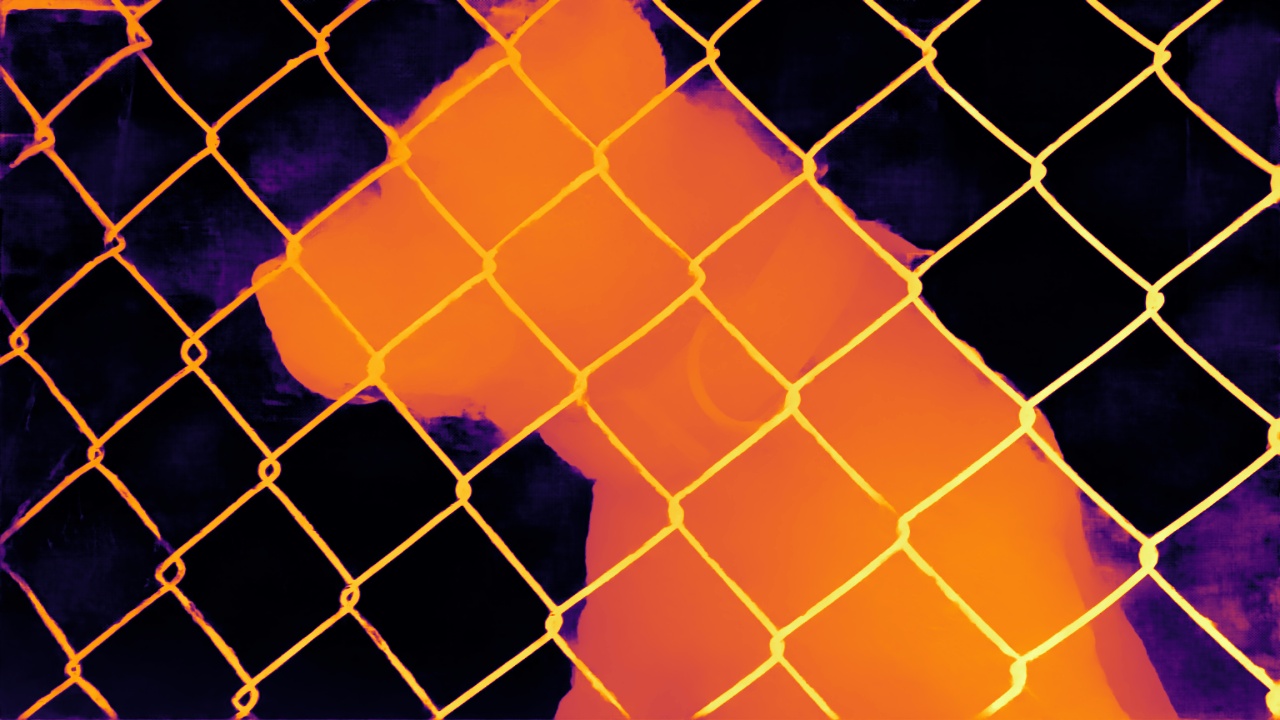}{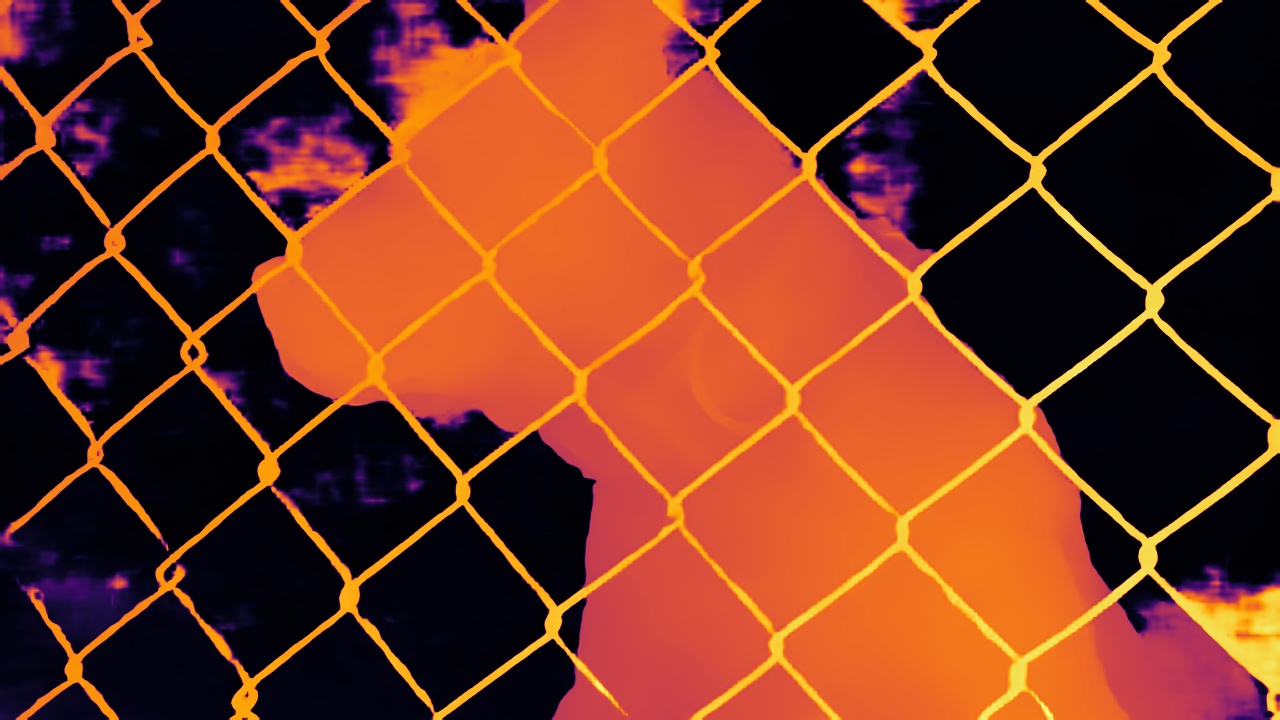}{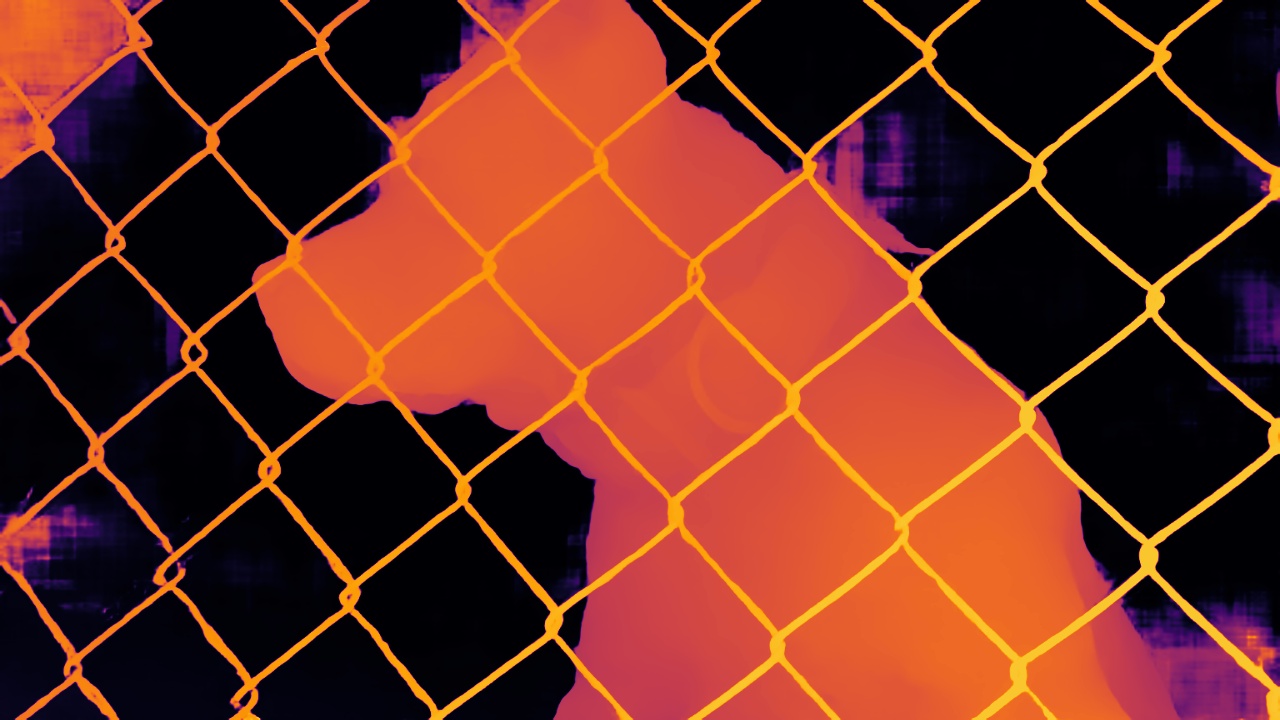}{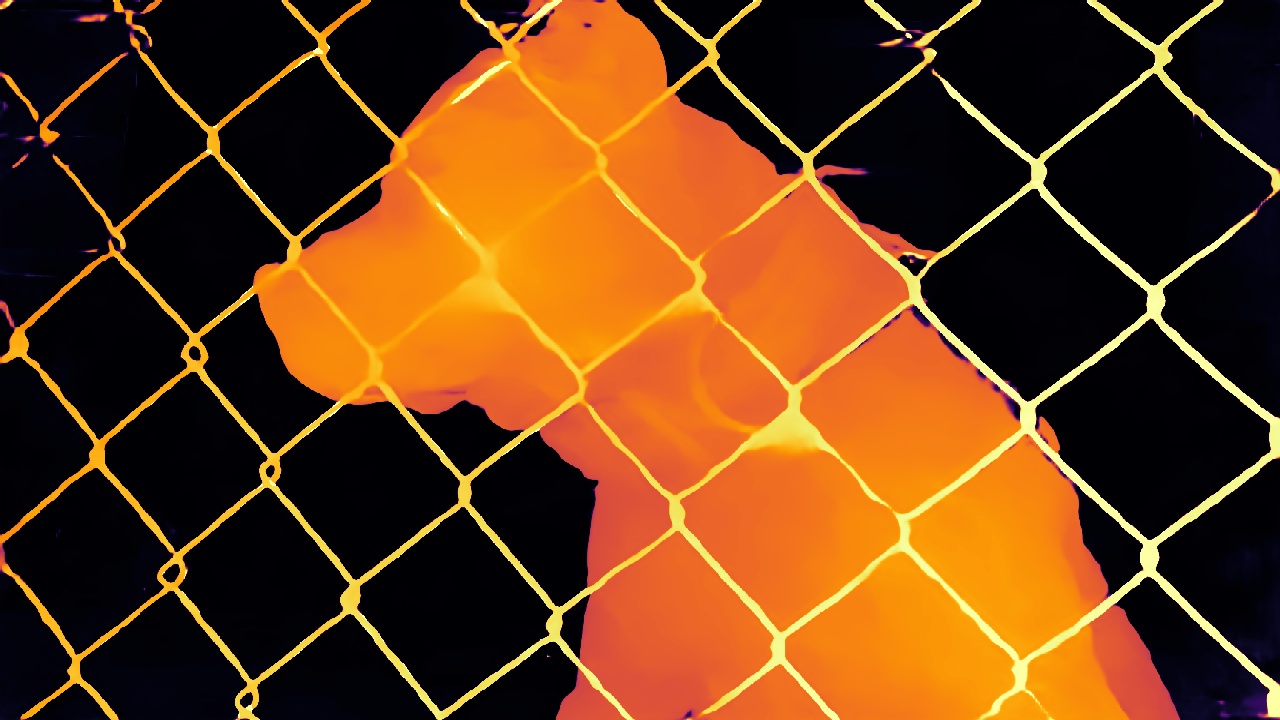}{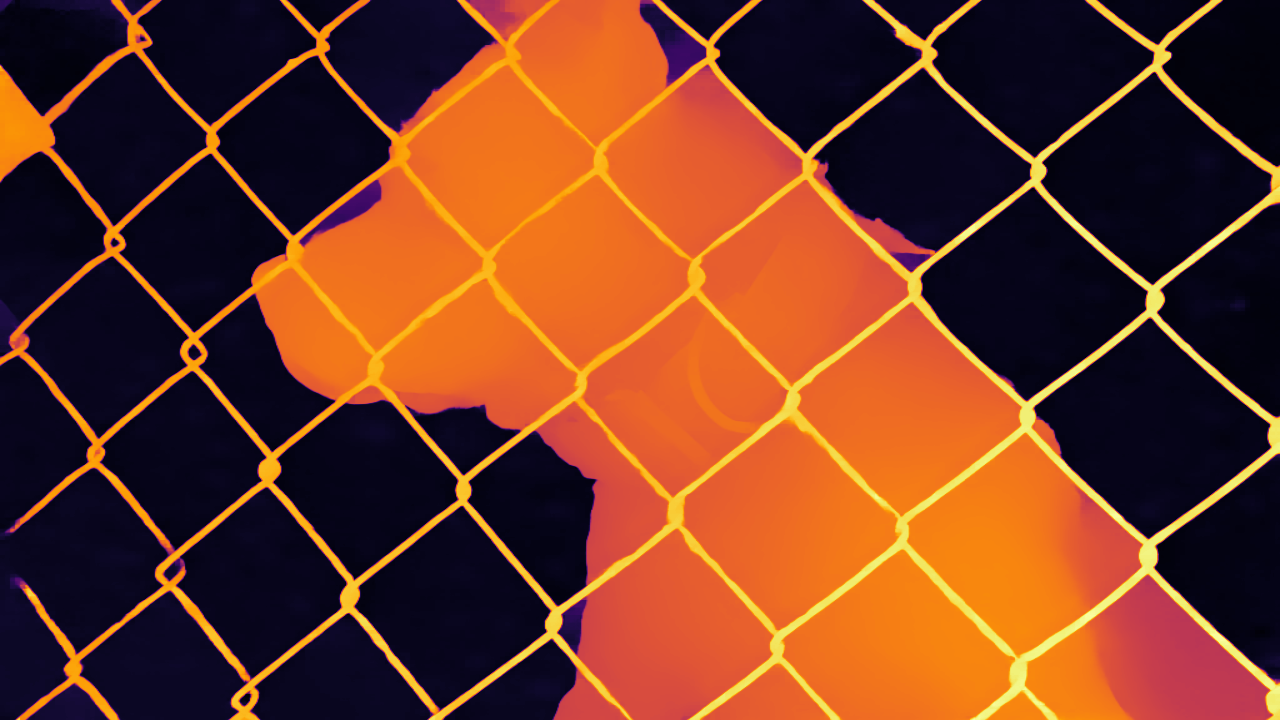}{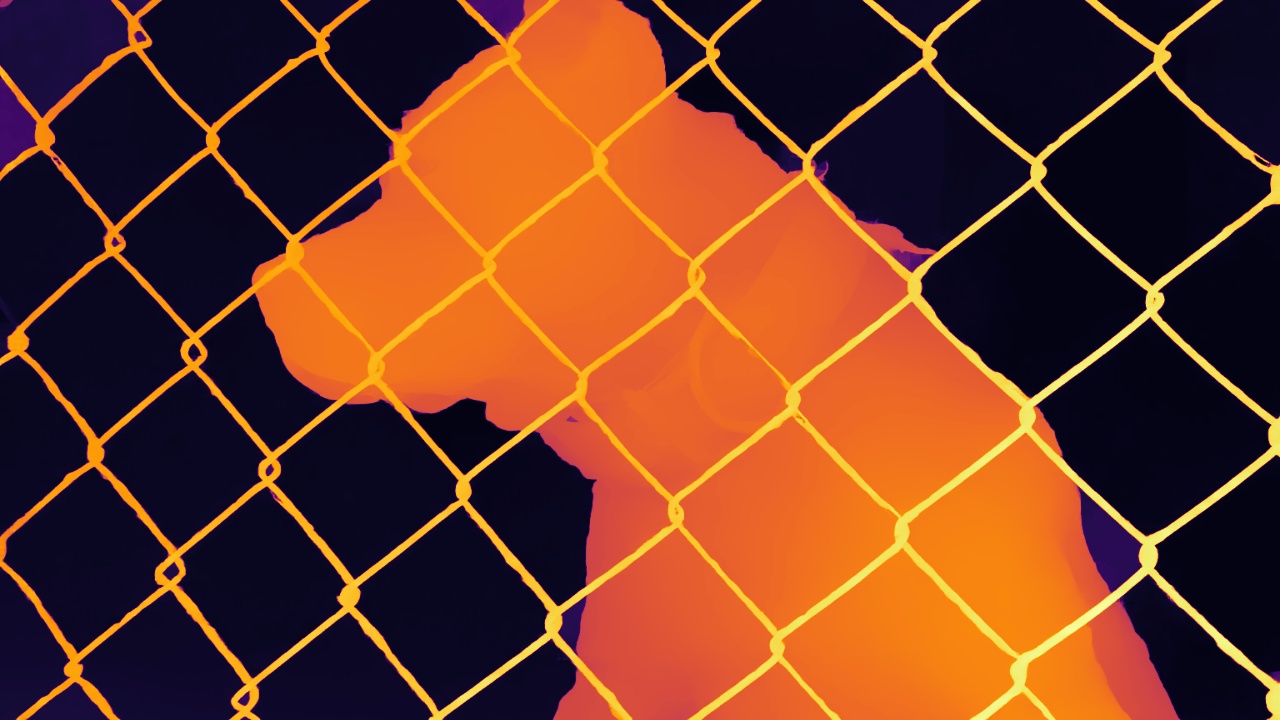}%
    \galleryRowCompare{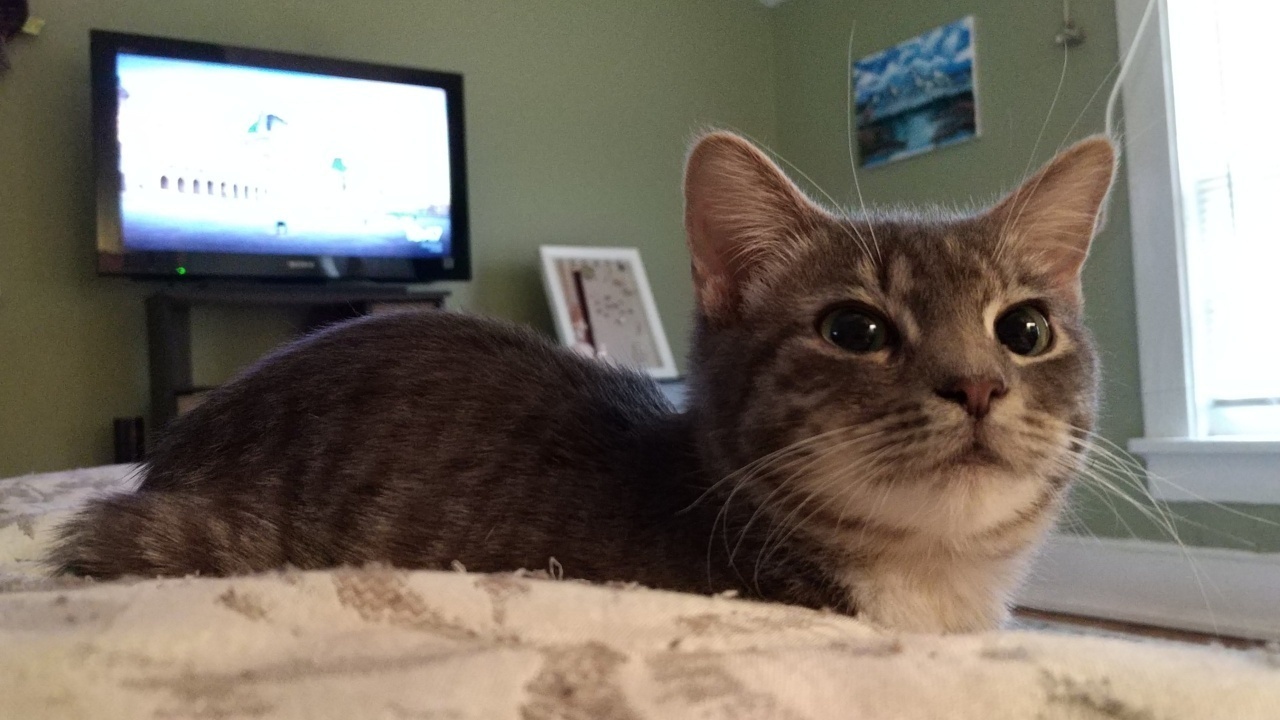}{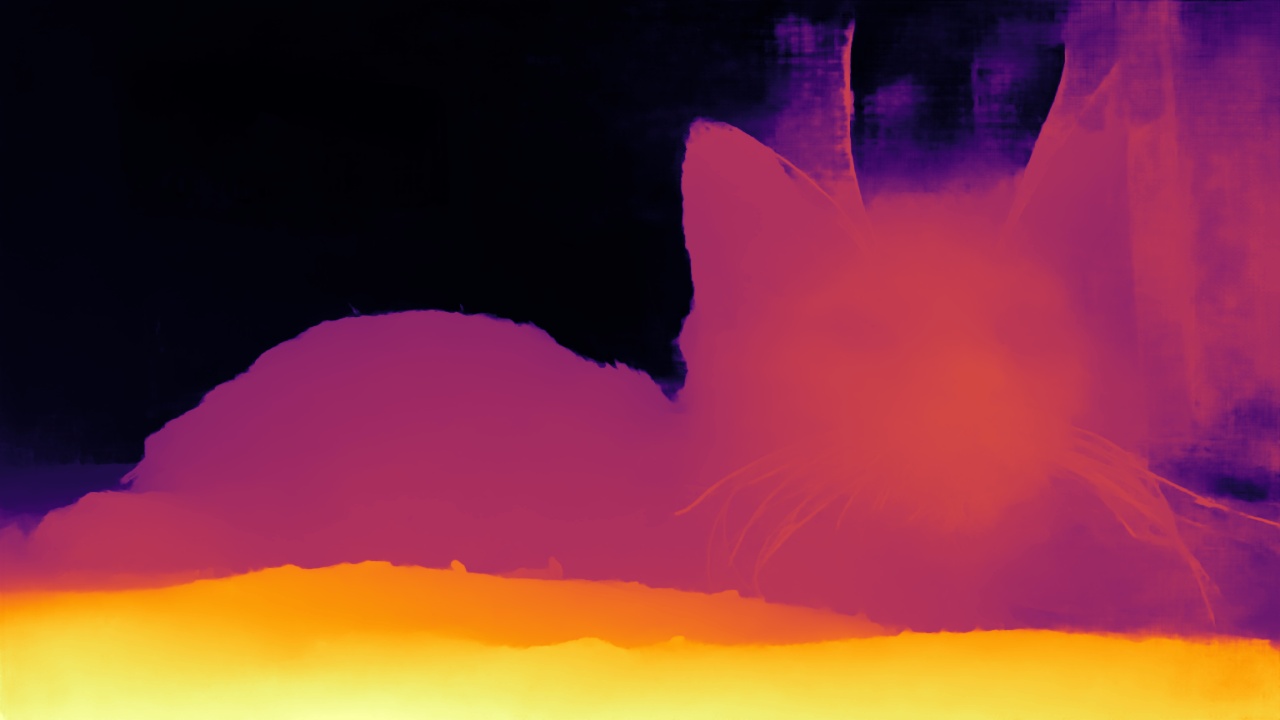}{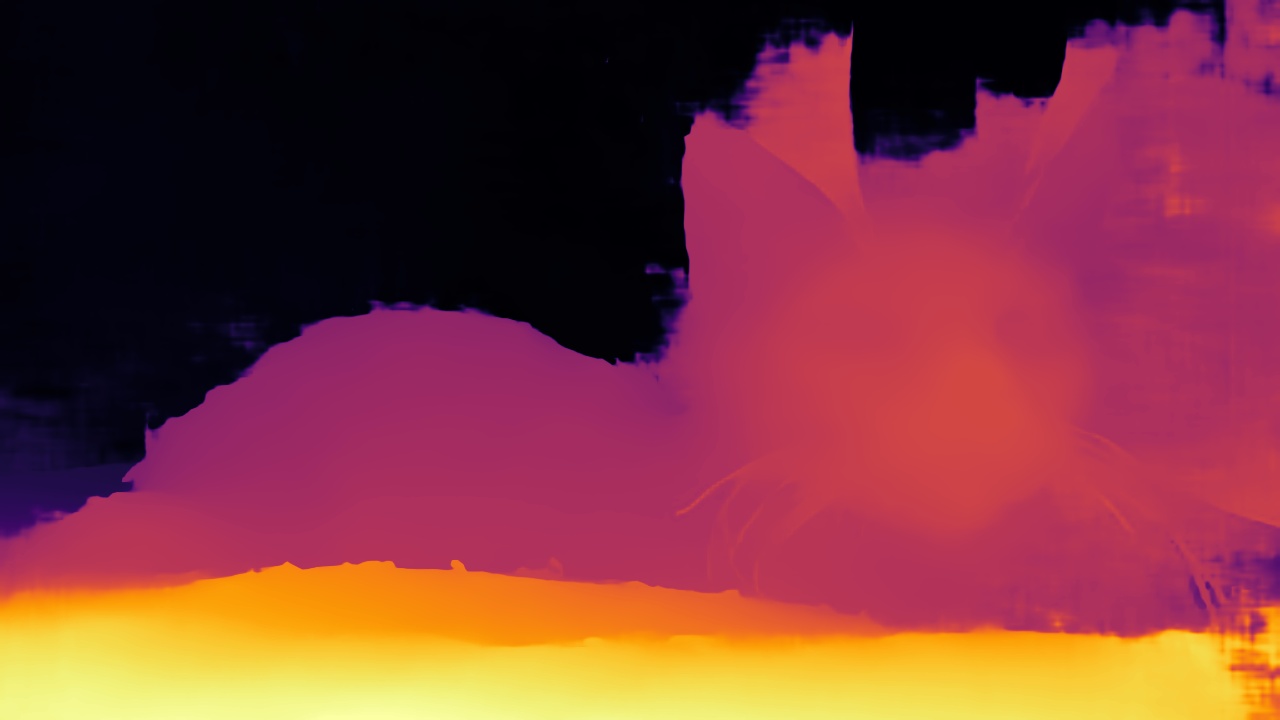}{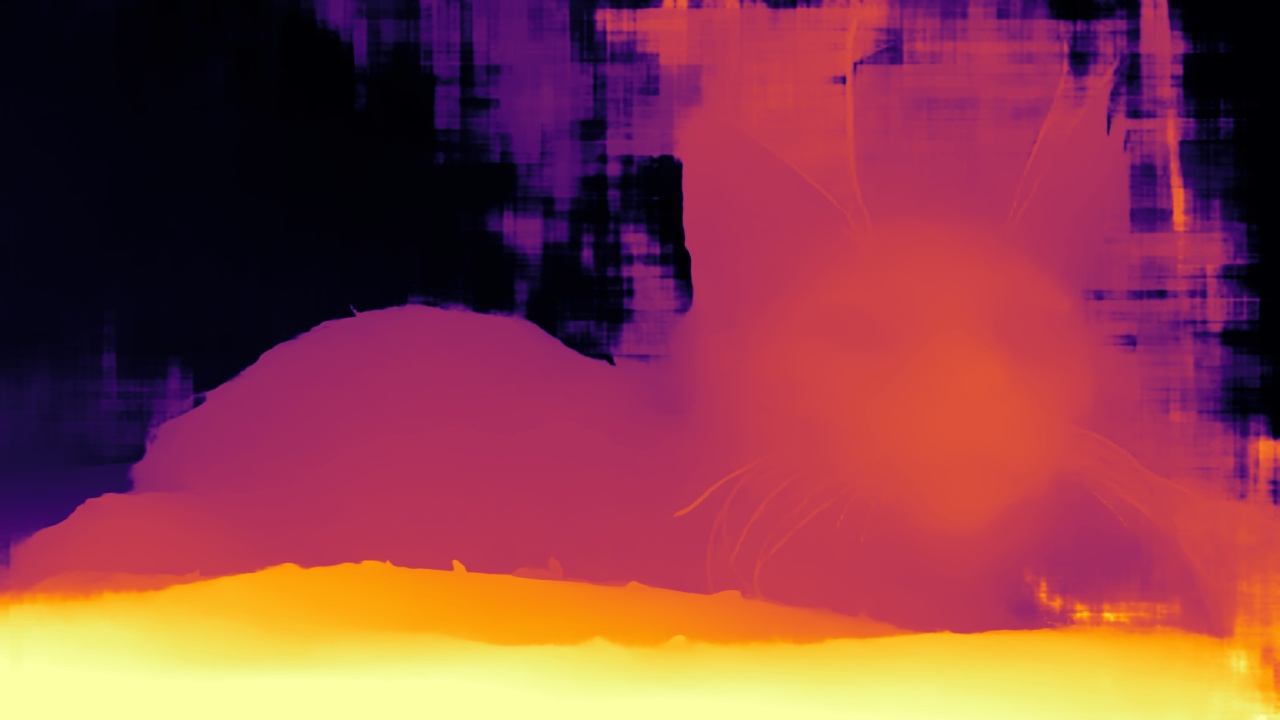}{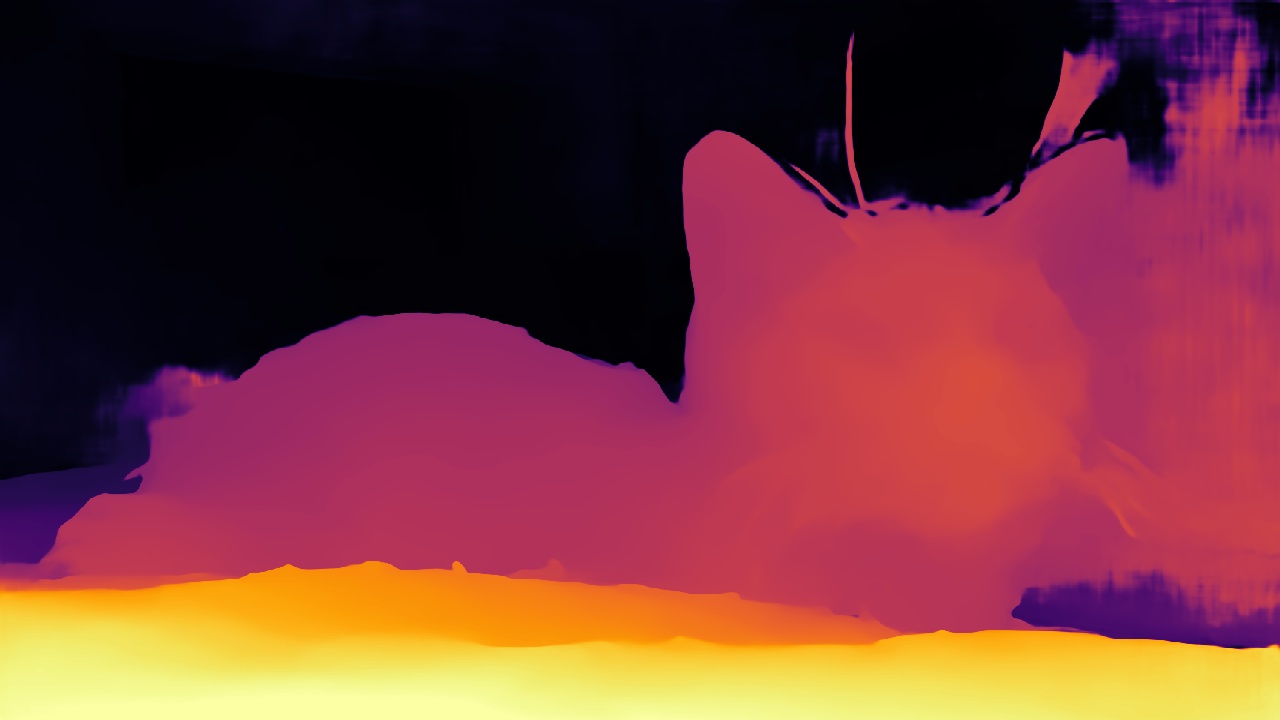}{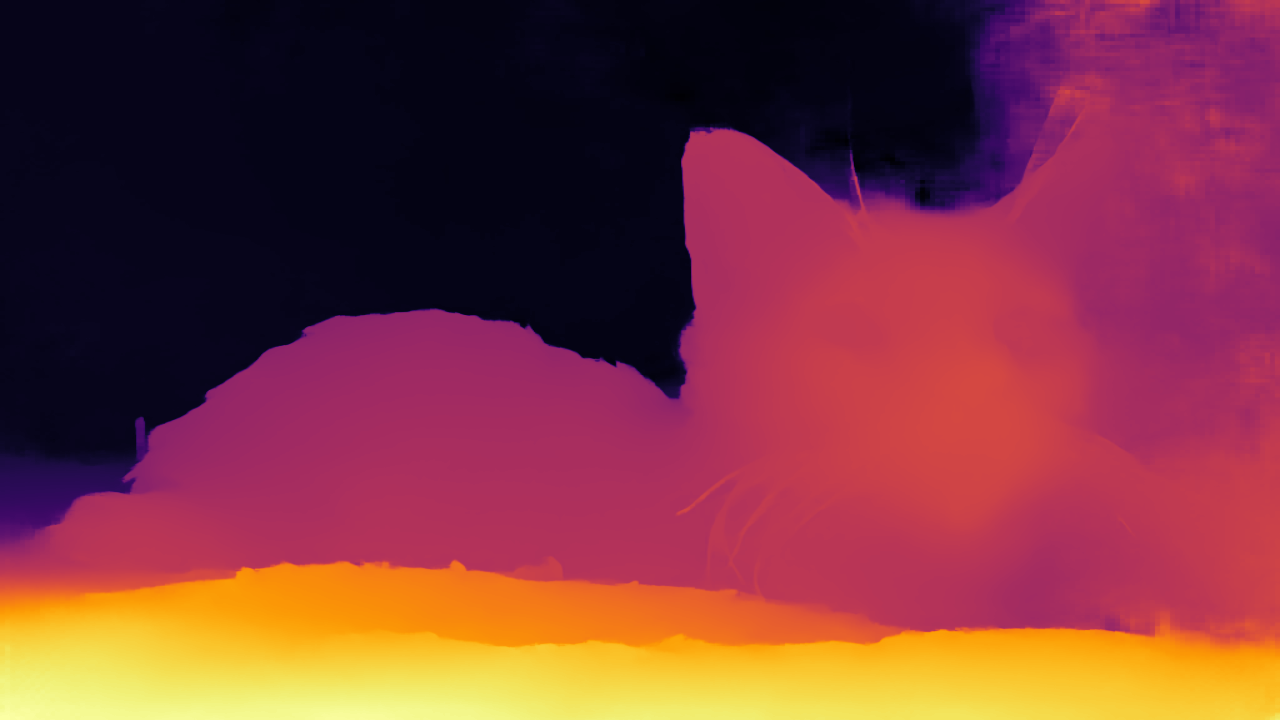}{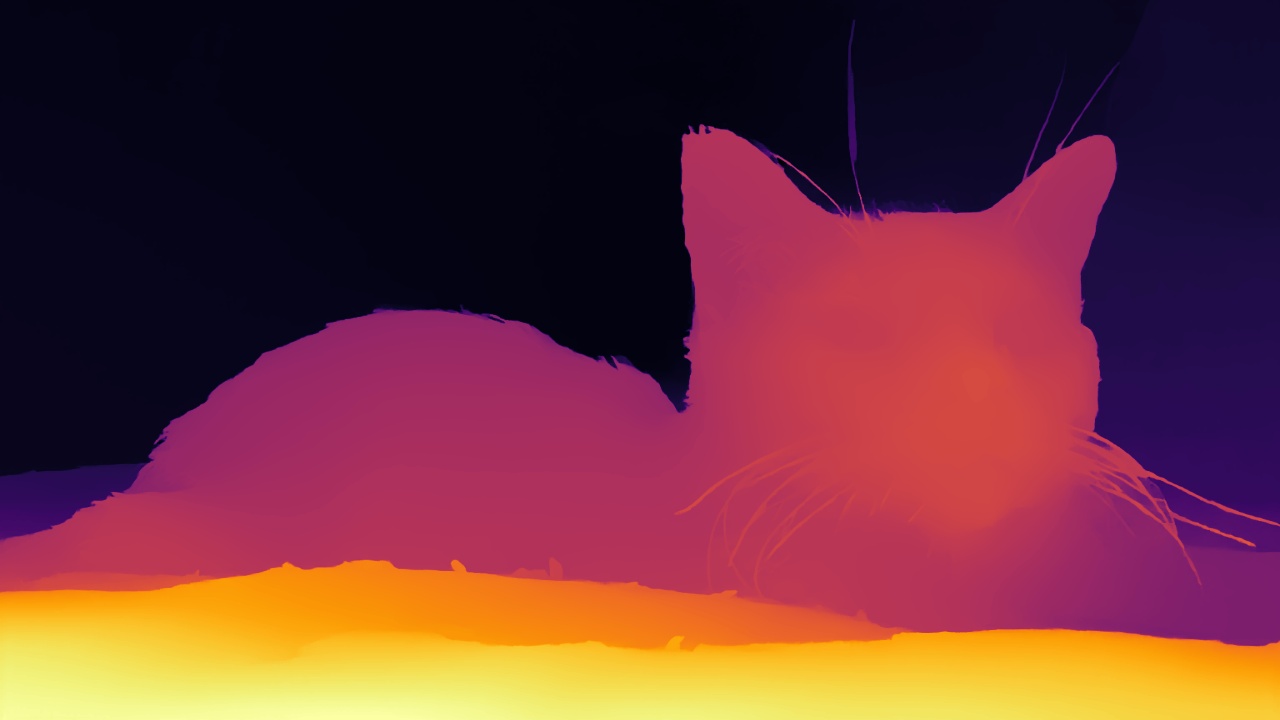}%
    \end{tabular}%
  }%
  \end{center}%
  \vspace{-1.5em}
  \caption{Comparison of results from  different methods on Holopix50K \cite{hua2020holopix50k} dataset. 
  Zoom in for best view. 
%   See the supplementary materials for more results.
  }
%   \vspace{-0.5em}
  \label{fig:holopix50k}
\end{figure*}

\subsection{Comparisons with State-of-the-art}

\textbf{Middlebury.}
We train our network on 23 pairs of images (including 13 additional pairs with ground truth) from Middlebury 2014 dataset together with our full training set without fine-tuning. The proportion of Middlebury training set is augmented to 2\% of the full training set. We evaluate the test set at $1536 \times 2048$ using resized full-resolution images where 2-stage inference is adopted, and the results are submitted to the online 
% leaderboard\footnote{\url{https://vision.middlebury.edu/stereo/eval3/}}
leaderboard
. We achieve the \nth{1} place on the majority of the metrics among more than 120 other methods, surpassing the published state-of-the-art by 21.73\% on the bad 2.0 metric, 31.00\% on the A95 metric. The quantitative comparison results with other methods are shown in Tab.~\ref{tab:middlebury}. 

\textbf{ETH3D.}
We train our network on the whole training set with a proportion of 2\% augmented training data from ETH3D low-res two-view stereo dataset. Without fine-tuning, we evaluate the test set at the size of $768 \times 1024$ where 2-stage inference is adopted. At the time of writing, we achieve state-of-the-art among published methods on the online
% benchmark\footnote{\url{https://www.eth3d.net/low\_res\_two\_view}} 
benchmark
for all metrics. Our method surpasses the published state-of-the-art by 59.84\% on the bad 1.0 metric. Quantitative comparisons are shown in Tab.~\ref{tab:eth3d}.

\def \rank#1{$^{\textcolor{red}{#1}}$ }
\begin{table}[t]
  \centering
  \small
  \setlength{\tabcolsep}{2.pt}
  \begin{tabular}{lccccc}
    \toprule
    Method & Bad 2.0 & Bad 1.0 & AvgErr & RMS & A95 \\
    \midrule
    CREStereo (Ours) & \textbf{3.71}\rank{1} & \textbf{8.25}\rank{1} & \textbf{1.15}\rank{1} & \textbf{7.70}\rank{1} & \textbf{1.58}\rank{1} \\
    RAFT-Stereo \cite{lipson2021raft} & \underline{4.74}\rank{2} & \underline{9.37}\rank{2} & \underline{1.27}\rank{2} & 8.41\rank{3} & \underline{2.29}\rank{2} \\
    LocalExp \cite{taniai2017continuous} & 5.43\rank{5} & 13.9\rank{10} & 2.24\rank{13} & 13.4\rank{23} & 4.81\rank{17} \\
    HITNet \cite{tankovich2021hitnet} & 6.46\rank{14} & 13.3\rank{4} & 1.71\rank{4} & 9.97\rank{5} & 4.26\rank{9} \\
    LEAStereo \cite{cheng2020hierarchical} & 7.15\rank{18} & 20.8\rank{40} & 1.43\rank{3} & \underline{8.11}\rank{2} & 2.65\rank{3} \\
    SDR \cite{yan2019segment} & 7.69\rank{24} & 18.8\rank{32} & 2.94\rank{32} & 15.4\rank{43} & 7.13\rank{30} \\
    MC-CNN-acrt \cite{zbontar2015computing} & 8.08\rank{27} & 17.1\rank{23} & 3.82\rank{58} & 21.3\rank{86} & 14.1\rank{55} \\
    CFNet \cite{shen2021cfnet} & 10.1\rank{37} & 19.6\rank{33} & 3.49\rank{46} & 15.4\rank{44} & 16.4\rank{58} \\
    HSMNet \cite{wang2020improving} & 10.2\rank{38} & 24.6\rank{48} & 2.07\rank{5} & 10.3\rank{8} & 4.32\rank{10} \\
    AdaStereo \cite{song2021adastereo} & 13.7\rank{59} & 29.5\rank{61} & 2.22\rank{10} & 10.2\rank{7} & 5.67\rank{25} \\
    AANet++ \cite{xu2020aanet} & 15.4\rank{66} & 25.5\rank{51} & 6.37\rank{94} & 23.5\rank{103} & 48.8\rank{112} \\
    % EdgeStereo \cite{song2020edgestereo} & 18.7\rank{74} & 32.4\rank{77} & 2.68\rank{26} & 9.84\rank{4} & 9.78\rank{42} \\
    % iResNet \cite{liang2018learning} & 22.9\rank{89} & 38.8\rank{92} & 3.31\rank{42} & 11.3\rank{12} & 12.5\rank{50} \\
    \bottomrule
  \end{tabular}
  \vspace{-5pt}
  \caption{Quantitative results on Middlebury benchmark.}
  \label{tab:middlebury}
%   \vspace{-10pt}
\end{table}

\def \rank#1{$^{\textcolor{red}{#1}}$ }
\begin{table}[t]
  \centering
  \small
  \setlength{\tabcolsep}{3.pt}
  \begin{tabular}{lccccc}
    \toprule
    Method & Bad 1.0 & Bad 0.5 & AvgErr & RMSE \\
    \midrule
    CREStereo (Ours) & \textbf{0.98}\rank{1} & \textbf{3.58}\rank{1} & \textbf{0.13}\rank{1} & \textbf{0.28}\rank{1} \\
    RAFT-Stereo \cite{lipson2021raft} & \underline{2.44}\rank{5} & \underline{7.04}\rank{4} & \underline{0.18}\rank{3} & \underline{0.36}\rank{3} \\
    HITNet \cite{tankovich2021hitnet} & 2.79\rank{9} & 7.83\rank{6} & 0.20\rank{6} & 0.46\rank{10} \\
    AdaStereo \cite{song2021adastereo} & 3.09\rank{12} & 10.22\rank{18} & 0.24\rank{14} & 0.44\rank{7} \\
    CFNet \cite{shen2021cfnet} & 3.31\rank{17} & 9.87\rank{15} & 0.24\rank{14} & 0.51\rank{19} \\
    GwcNet \cite{guo2019group} & 3.66\rank{25} & 12.04\rank{37} & 0.29\rank{40} & 0.67\rank{52} \\
    iResNet \cite{liang2018learning} & 3.68\rank{26} & 10.26\rank{19} & 0.24\rank{14} & 0.51\rank{19} \\
    HSMNet \cite{wang2020improving} & 4.00\rank{36} & 11.33\rank{28} & 0.28\rank{36} & 0.62\rank{43} \\
    AANet \cite{xu2020aanet} & 5.01\rank{52} & 13.16\rank{45} & 0.31\rank{45} & 0.68\rank{57} \\
    GANet \cite{zhang2019ga} & 6.56\rank{67} & 25.41\rank{108} & 0.43\rank{83} & 0.75\rank{73} \\ 
    %% EdgeStereo \cite{song2020edgestereo} & 6.76\rank{71} & 18.75\rank{79} & 0.39\rank{76} & 0.81\rank{82} \\
    \bottomrule
  \end{tabular}
  \vspace{-5pt}
  \caption{Quantitative results on ETH3D benchmark.}
  \label{tab:eth3d}
\end{table}

\textbf{KITTI.}
Different from the training procedure for Middlebury and ETH3D, we fine-tune the model pre-trained on the full training set for another 50K iterations on KITTI 2012 and 2015 training sets. The initial learning rate is set to 0.0001. We augment the proportion of KITTI datasets to 75\% with the rest part randomly sampled from the whole training set. During evaluation, we pad the input to $384 \times 1248$ before feeding to the network and single stage inference is adopted. We achieve competitive performance on both
% datasets\footnote{\url{http://www.cvlibs.net/datasets/kitti/eval_stereo.php}}
datasets
, surpassing LEAStereo \cite{cheng2020hierarchical} in KITTI 2012 by 9.47\% on Out-Noc under 2 pixels error threshold. We show a visual comparison of KITTI 2015 in Fig.~\ref{fig:compare_kitti15}.

\subsection{Practical Performance}

Compared with the real-world images from standard stereo datasets which is limited in numbers and scenes, images taken from consumer-level devices pose greater challenges to stereo matching. For fair comparisons, we trained all other stereo networks with author-released code and recommended settings on our full training set. 

\textbf{Holopix50K.}
Fig.~\ref{fig:holopix50k} shows the qualitative comparison results of our network with several published stereo matching on Holopix50K \cite{hua2020holopix50k} dataset in varied scenes. Pre-rectification were performed to eliminate possible negative disparity. The visual results show that our method has a significant advantage in thin objects like cat whiskers and wire meshes. We also achieve better performance on textureless areas like walls and windows. 

\begin{figure}
    \centering
    \includegraphics[width=0.9\linewidth]{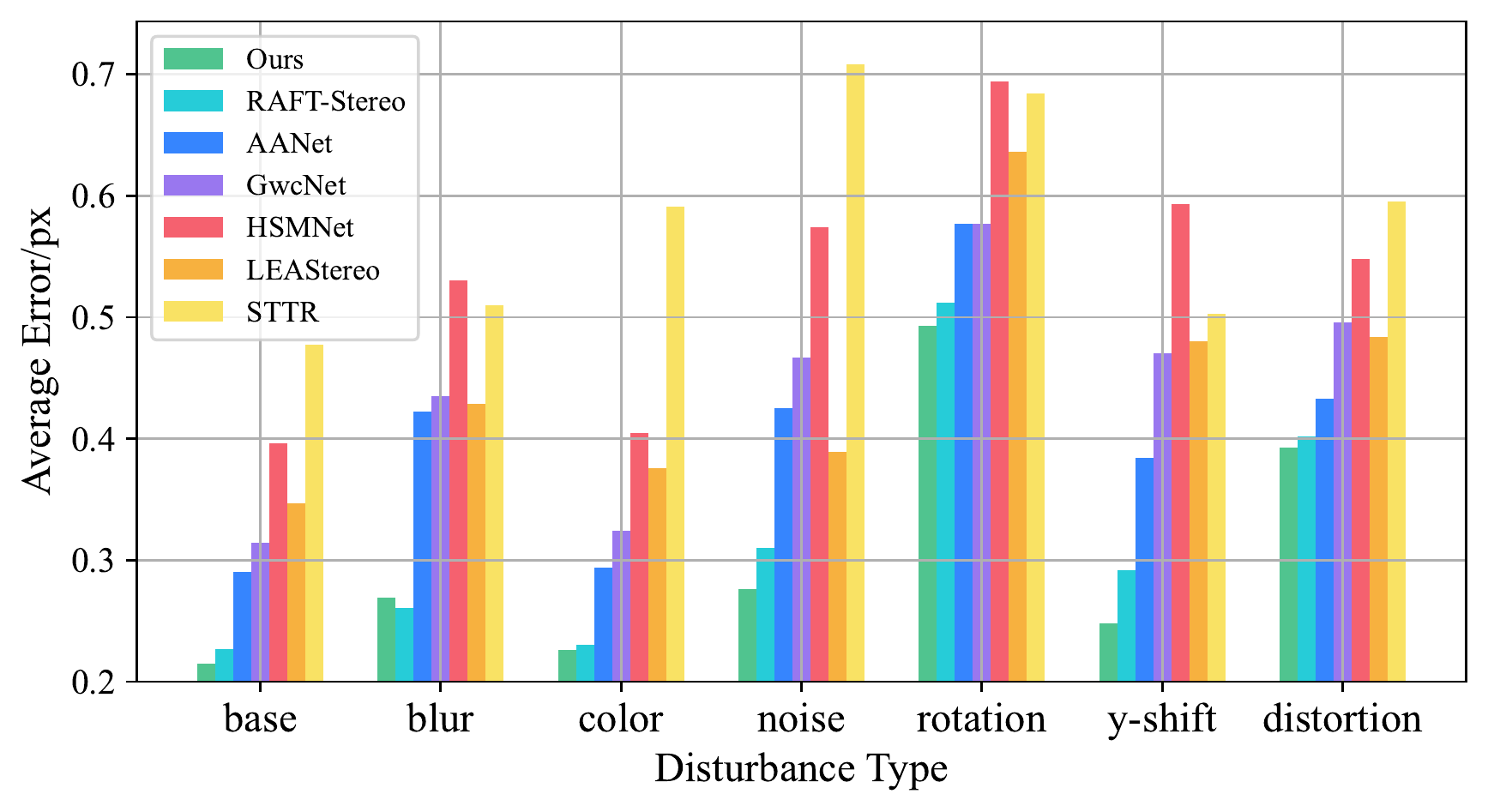}
    \vspace{-0.5em}
    \caption{Comparisons among methods on the ETH3D training dataset with different types of disturbances.}
    \label{fig:disturb_avgerr_noccl}
    \vspace{-0.5em}
\end{figure}

\textbf{Disturbed ETH3D.}
We simulate common disturbances in practical scenes on ETH3D dataset to test the robustness of our proposed method and list the quantitative results in Fig.~\ref{fig:disturb_avgerr_noccl}. The disturbances here include image blur, color transform, chromatic noise, image perspective transform, vertical shift and spatial distortion. The results demonstrate that our method is less vulnerable to these disturbances.

\begin{table}[t]
  \centering
    \small
    \setlength{\tabcolsep}{3.pt}
    \begin{tabular}{lccc}
      \toprule
      Method & mxIoU & mxIoUbd \\
      \midrule
      Ours & {\bf 97.50\%} & {\bf 72.61\%} \\
      RAFT-Stereo \cite{lipson2021raft} & \underline{94.58\%} & \underline{69.26\%} \\
      HSMNet \cite{wang2020improving} & 91.70\% & 60.17\% \\
      AANet \cite{xu2020aanet} & 91.02\% & 63.70\% \\
      GwcNet \cite{guo2019group} & 90.77\% & 64.26\% \\
      STTR \cite{li2020revisiting} & 90.82\% & 62.12\% \\
      LEAStereo \cite{cheng2020hierarchical} & 92.38\% & 58.06\% \\
      \bottomrule
    \end{tabular}
    \vspace{-0.5em}
  \caption{Quantitative results on 400 smartphone captured scenes. We choose the resolution with best performance for every method. }
  \label{tab:maxiou}
\end{table}

\begin{figure}
    \centering
    \begin{subfigure}[b]{0.24\linewidth}
        \includegraphics[width=\linewidth]{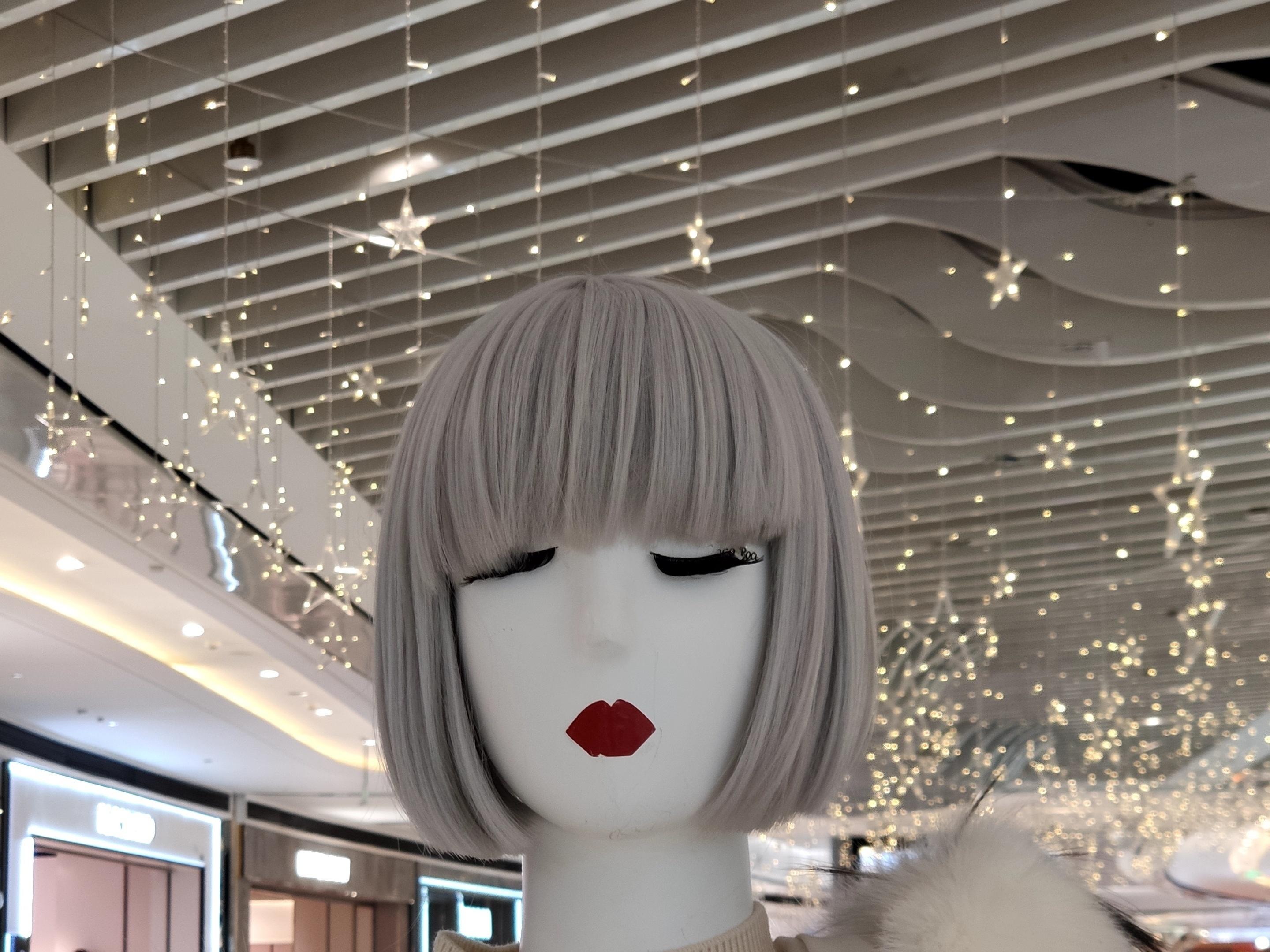}
    \end{subfigure}
    \begin{subfigure}[b]{0.24\linewidth}
        \includegraphics[width=\linewidth]{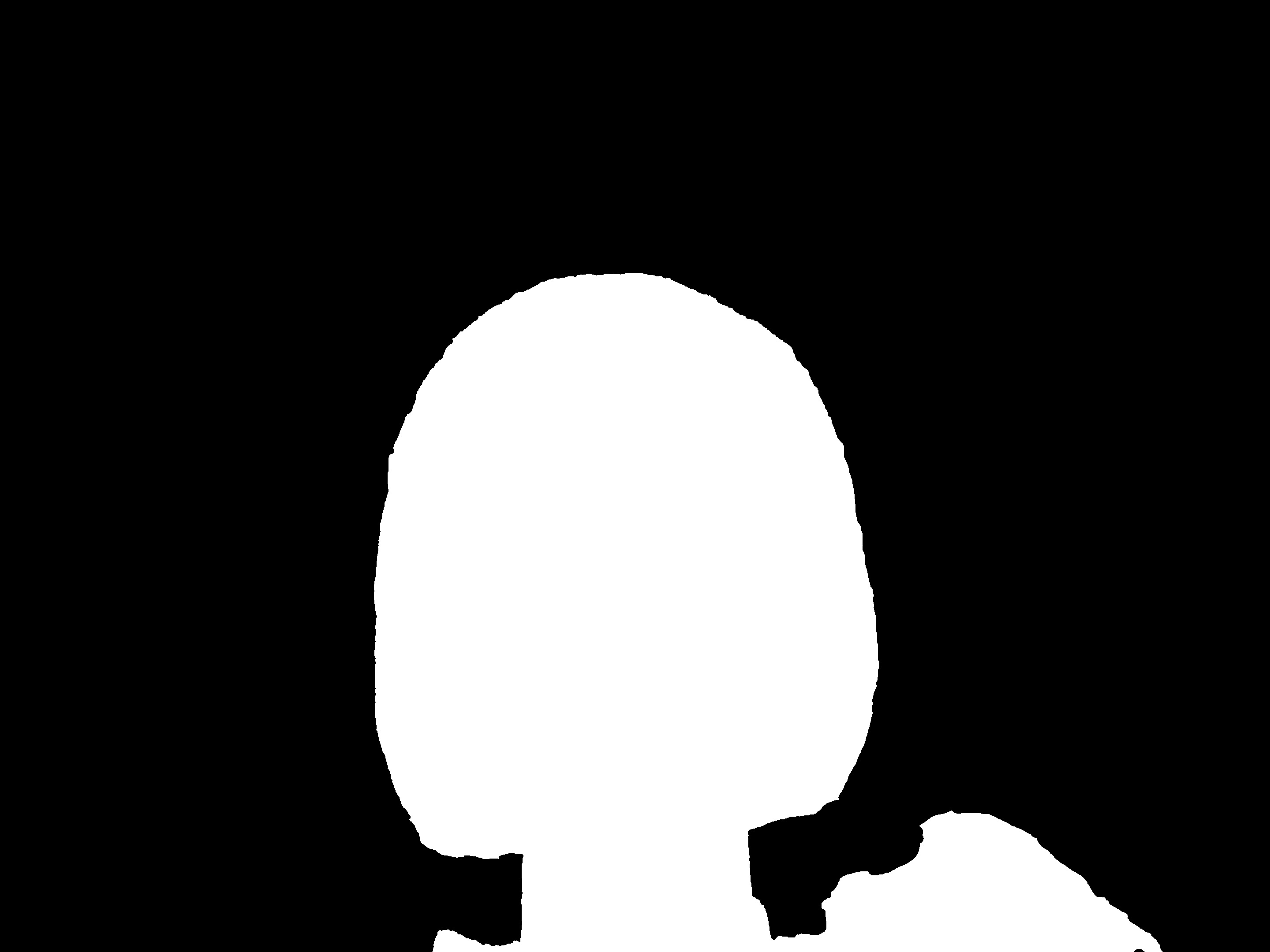}
    \end{subfigure}
    \begin{subfigure}[b]{0.24\linewidth}
        \includegraphics[width=\linewidth]{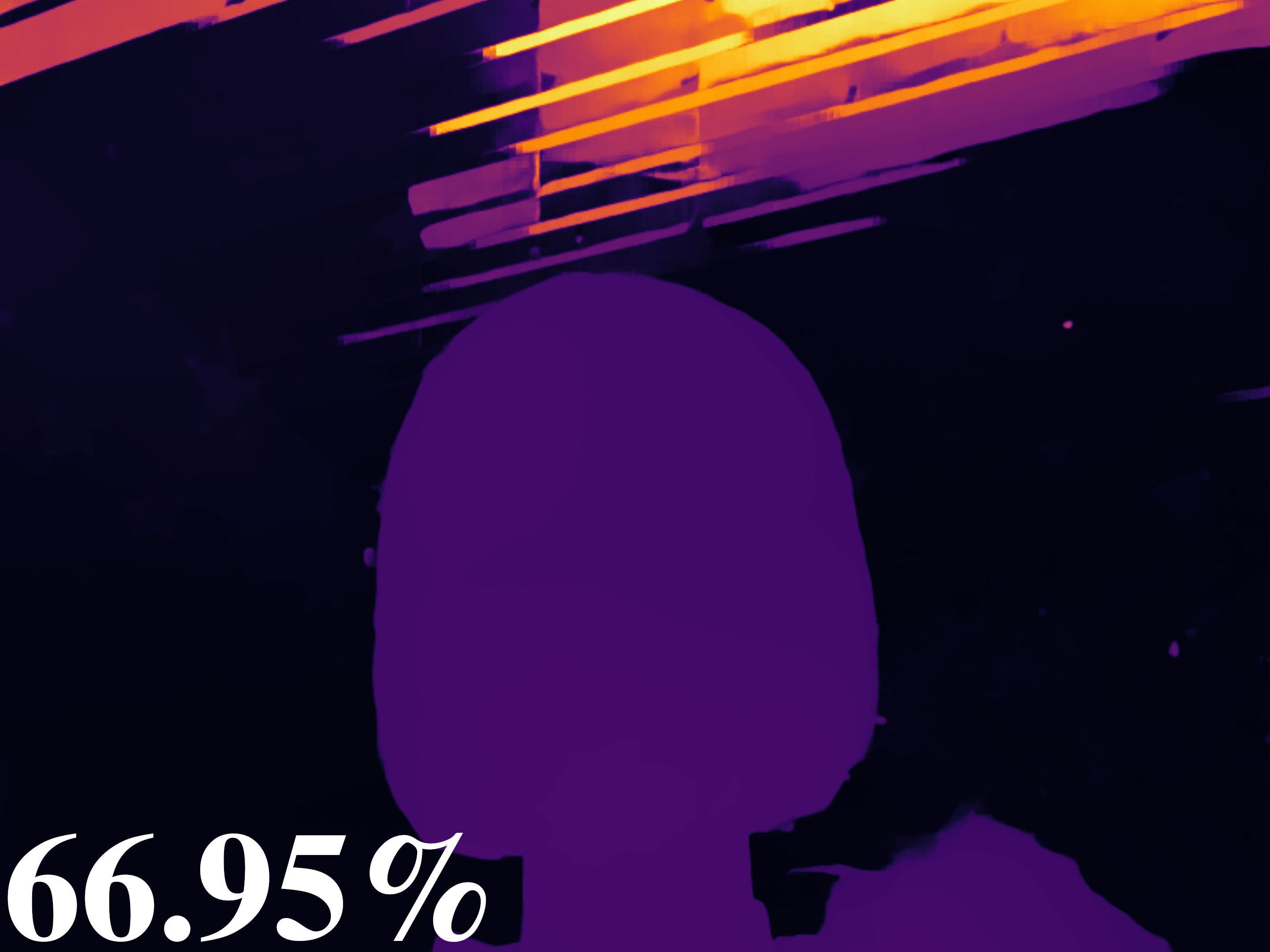}
    \end{subfigure}
    \begin{subfigure}[b]{0.24\linewidth}
        \includegraphics[width=\linewidth]{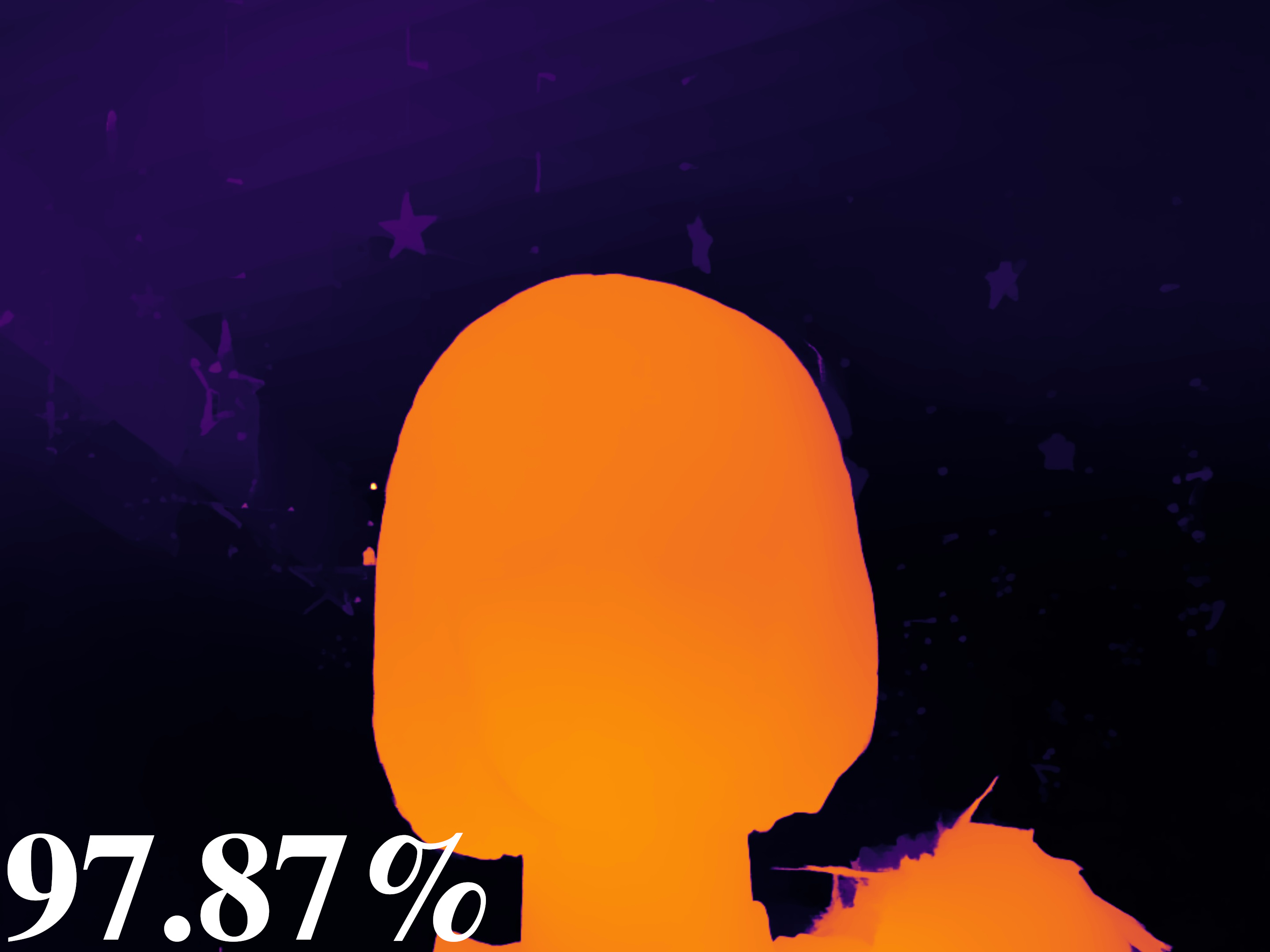}
    \end{subfigure}
    \\
    \vspace{0.15em}
    \begin{subfigure}[b]{0.24\linewidth}
      \includegraphics[width=\linewidth]{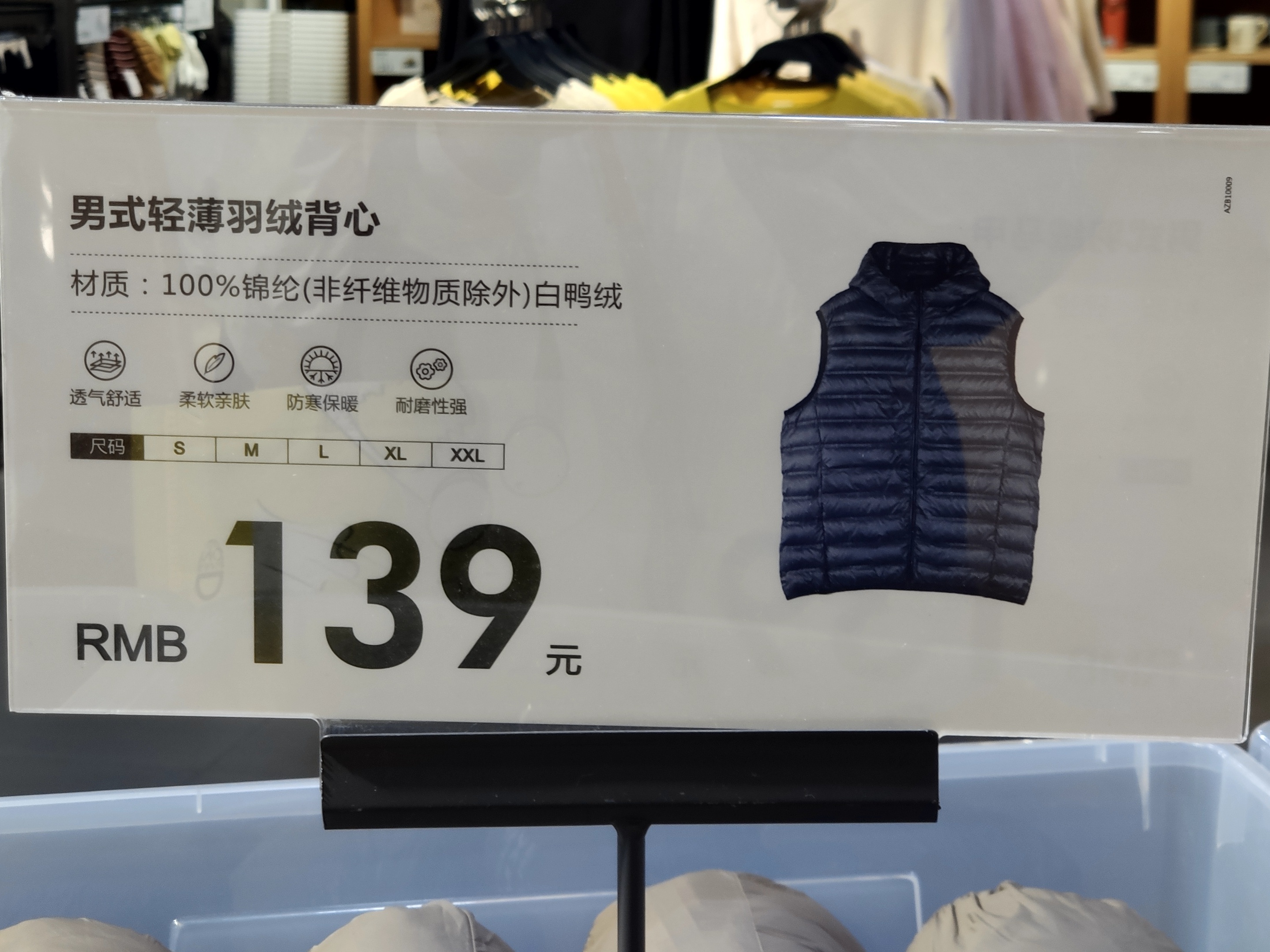}
      \caption{Left image}
    \end{subfigure}
    \begin{subfigure}[b]{0.24\linewidth}
      \includegraphics[width=\linewidth]{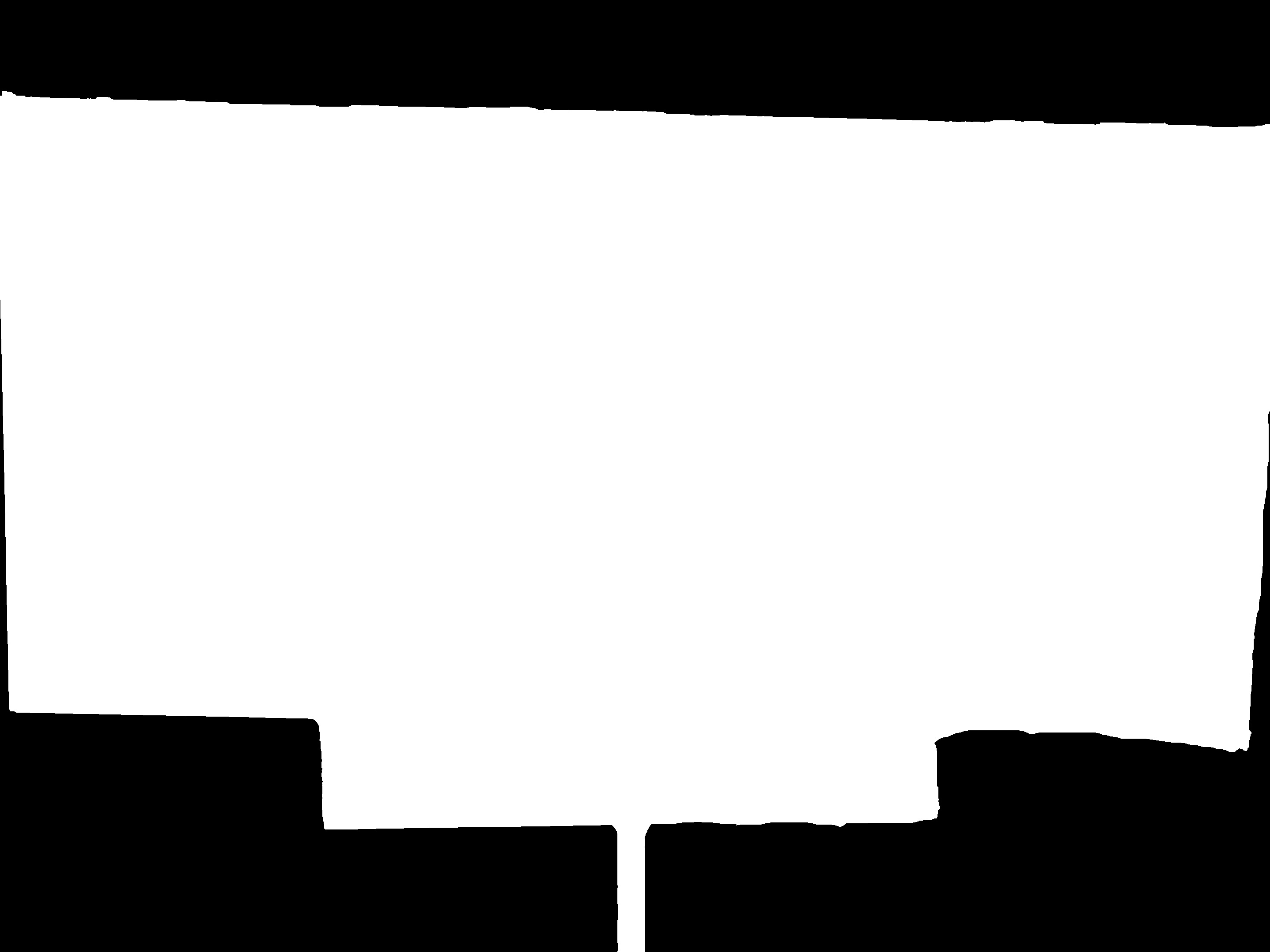}
      \caption{Mask}
    \end{subfigure}
    \begin{subfigure}[b]{0.24\linewidth}
      \includegraphics[width=\linewidth]{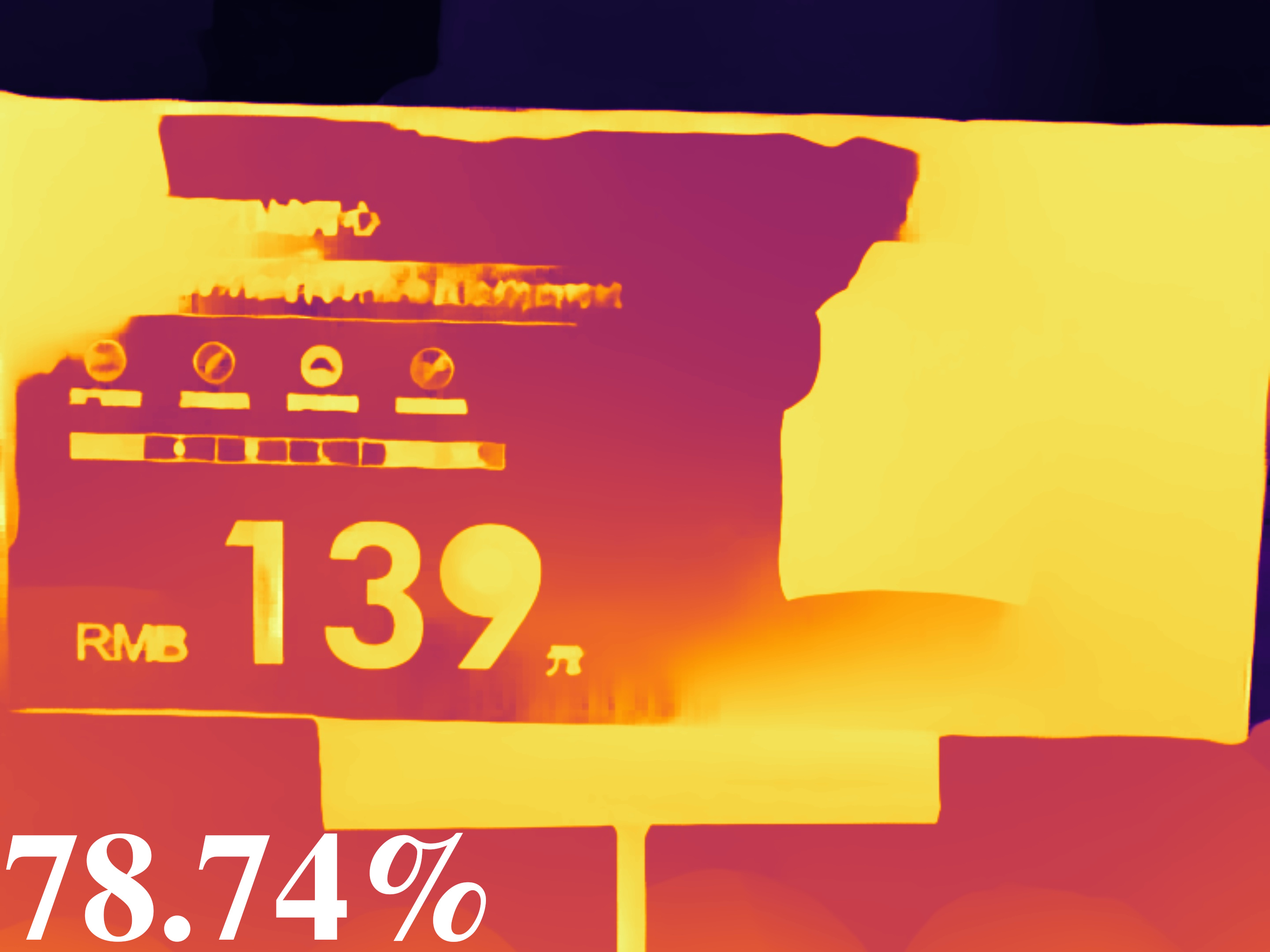}
      \caption{RAFT-Stereo}
    \end{subfigure}
    \begin{subfigure}[b]{0.24\linewidth}
      \includegraphics[width=\linewidth]{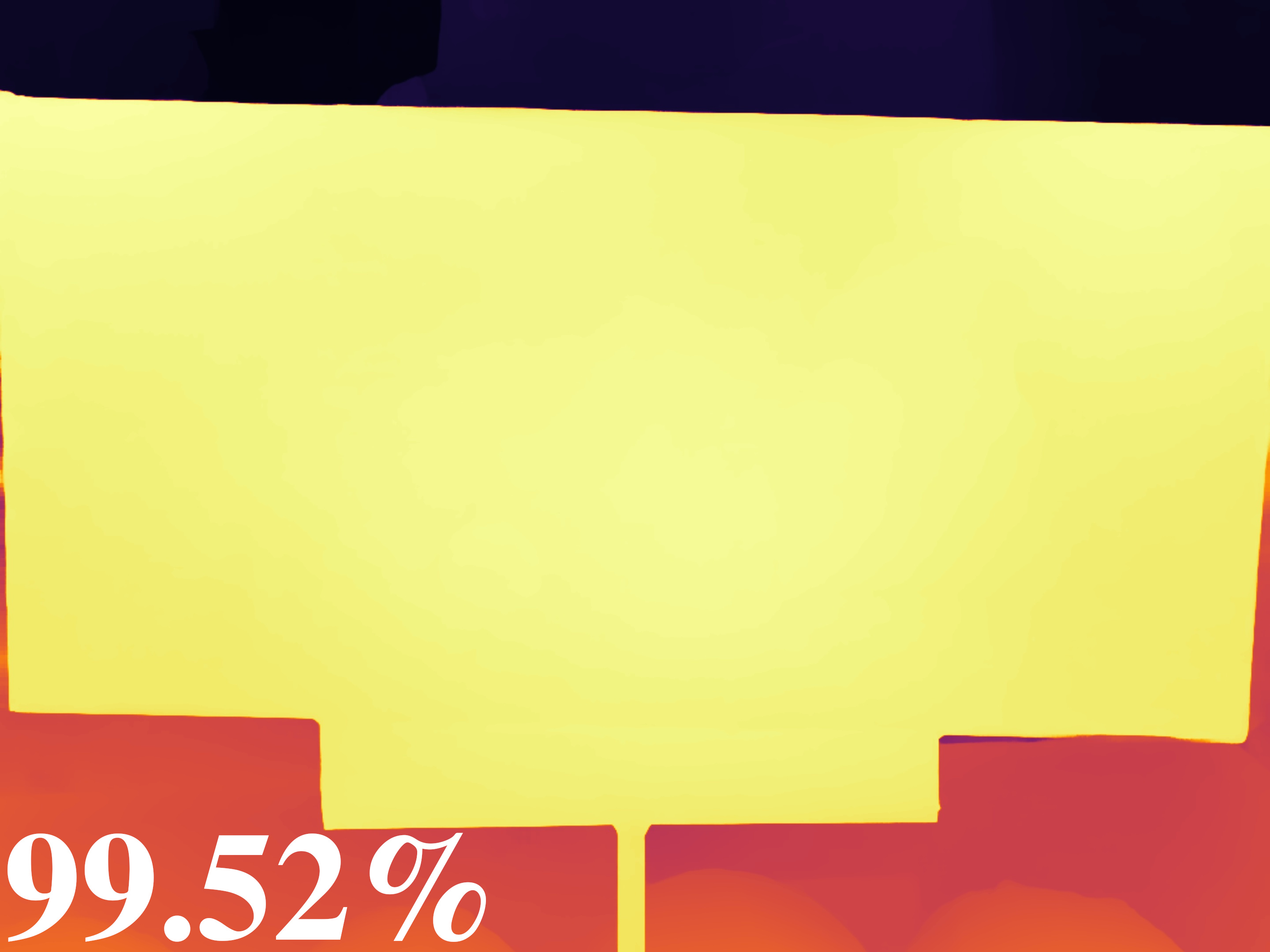}
      \caption{Ours}
    \end{subfigure}
    \\
    \vspace{-0.5em}
    \caption{Comparison of predicted disparities of repetitive-texture and  non-texture cases in smartphone photos with RAFT-Stereo \cite{lipson2021raft}. The mxIoU scores are marked in the figure.}
    \label{fig:smartphone}
\end{figure}

\textbf{Smartphone photos.}
Because it is difficult to obtain ground truth disparity in real-world scenes, an empirical way is to manually label a foreground mask $M_f$ for evaluating the disparity quality \cite{luo2020wavelet}. The metric of IoU (intersection over union) is commonly used in segmentation tasks. For a disparity map, we can place a threshold $t$ to obtain a foreground mask $M_t$, where the disparity values of the foreground are larger than $t$. The ``mxIoU" means the maximum IoU between $M_f$ and $M_t$ by changing $t$. Similarly, ``mxIoUbd" means mxIoU in a banded area within $p$ (we set $p=4$) pixels from the boundary of $M_f$. The quantitative and qualitative comparison results are shown in Tab.~\ref{tab:maxiou} and Fig.~\ref{fig:smartphone} respectively.

% \vspace{-5mm}

\section{Conclusion}
Despite unprecedented success of deep stereo networks, obstacles remain for accurately recovering disparities in real-world scenes. In this paper, we have presented CREStereo, a novel stereo matching network that attains state-of-the-art results on both public benchmarks and real-world scenes. Our key message here is that, both network architecture and training data deserve rigorous thoughts in order for an algorithm to truly work in the real world. Via cascaded recurrent network with adaptive correlation, we are able to recover delicate depth details better than existing methods; and we manage to better handle hard-case scenes like non-texture or repetitive-texture areas through careful design of our synthetic dataset. A limitation of our method is that the model is not yet efficient enough to run in current mobile applications. Future improvements could be made to adapt our network for various portable devices, preferably in real-time.

\section*{Acknowledgement}
We thank all the reviewers for their valuable comments. This work was supported by the National Natural Science Foundation of China (NSFC) under grants
No.61872067.

%%%%%%%%% REFERENCES
{\small
\bibliographystyle{ieee_fullname}
\bibliography{egbib}

\begin{thebibliography}{10}\itemsep=-1pt

\bibitem{bao2020instereo2k}
Wei Bao, Wei Wang, Yuhua Xu, Yulan Guo, Siyu Hong, and Xiaohu Zhang.
\newblock Instereo2k: a large real dataset for stereo matching in indoor
  scenes.
\newblock {\em Science China Information Sciences}, 63(11):1--11, 2020.

\bibitem{birchfield1999depth}
Stan Birchfield and Carlo Tomasi.
\newblock Depth discontinuities by pixel-to-pixel stereo.
\newblock {\em {International Journal of Computer Vision}}, 35(3):269--293,
  1999.

\bibitem{blender}
{Blender Online Community}.
\newblock {\em Blender - a 3D modelling and rendering package}.
\newblock Blender Foundation, Blender Institute, Amsterdam, 2021.

\bibitem{boykov2001fast}
Yuri Boykov, Olga Veksler, and Ramin Zabih.
\newblock Fast approximate energy minimization via graph cuts.
\newblock {\em {IEEE Trans. on Pattern Analysis and Machine Intelligence}},
  23(11):1222--1239, 2001.

\bibitem{sintel}
D.~J. Butler, J. Wulff, G.~B. Stanley, and M.~J. Black.
\newblock A naturalistic open source movie for optical flow evaluation.
\newblock In {\em {Proc. ECCV}}, pages 611--625, 2012.

\bibitem{shapenet}
Angel~X Chang, Thomas Funkhouser, Leonidas Guibas, Pat Hanrahan, Qixing Huang,
  Zimo Li, Silvio Savarese, Manolis Savva, Shuran Song, Hao Su, et~al.
\newblock Shapenet: An information-rich 3d model repository.
\newblock {\em arXiv preprint arXiv:1512.03012}, 2015.

\bibitem{chang2018pyramid}
Jia-Ren Chang and Yong-Sheng Chen.
\newblock Pyramid stereo matching network.
\newblock In {\em {Proc. CVPR}}, pages 5410--5418, 2018.

\bibitem{cheng2020hierarchical}
Xuelian Cheng, Yiran Zhong, Mehrtash Harandi, Yuchao Dai, Xiaojun Chang, Tom
  Drummond, Hongdong Li, and Zongyuan Ge.
\newblock Hierarchical neural architecture search for deep stereo matching.
\newblock {\em arXiv preprint arXiv:2010.13501}, 2020.

\bibitem{deschaud2021kitti}
Jean-Emmanuel Deschaud.
\newblock Kitti-carla: a kitti-like dataset generated by carla simulator.
\newblock {\em arXiv preprint arXiv:2109.00892}, 2021.

\bibitem{flownet}
Alexey Dosovitskiy, Philipp Fischer, Eddy Ilg, Philip Häusser, Caner Hazirbas,
  Vladimir Golkov, Patrick van~der Smagt, Daniel Cremers, and Thomas Brox.
\newblock Flownet: Learning optical flow with convolutional networks.
\newblock In {\em 2015 IEEE International Conference on Computer Vision
  (ICCV)}, pages 2758--2766, 2015.

\bibitem{kitti}
Andreas Geiger, Philip Lenz, and Raquel Urtasun.
\newblock Are we ready for autonomous driving? the kitti vision benchmark
  suite.
\newblock In {\em {Proc. CVPR}}, pages 3354--3361, 2012.

\bibitem{guo2019group}
Xiaoyang Guo, Kai Yang, Wukui Yang, Xiaogang Wang, and Hongsheng Li.
\newblock Group-wise correlation stereo network.
\newblock In {\em {Proc. CVPR}}, pages 3273--3282, 2019.

\bibitem{he2021semi}
Ju He, Enyu Zhou, Liusheng Sun, Fei Lei, Chenyang Liu, and Wenxiu Sun.
\newblock Semi-synthesis: A fast way to produce effective datasets for stereo
  matching.
\newblock In {\em {Proc. CVPR}}, pages 2884--2893, 2021.

\bibitem{hirschmuller2005accurate}
Heiko Hirschmuller.
\newblock Accurate and efficient stereo processing by semi-global matching and
  mutual information.
\newblock In {\em {Proc. CVPR}}, volume~2, pages 807--814, 2005.

\bibitem{hirschmuller2002real}
Heiko Hirschm{\"u}ller, Peter~R Innocent, and Jon Garibaldi.
\newblock Real-time correlation-based stereo vision with reduced border errors.
\newblock {\em {International Journal of Computer Vision}}, 47(1):229--246,
  2002.

\bibitem{hua2020holopix50k}
Yiwen Hua, Puneet Kohli, Pritish Uplavikar, Anand Ravi, Saravana Gunaseelan,
  Jason Orozco, and Edward Li.
\newblock Holopix50k: A large-scale in-the-wild stereo image dataset.
\newblock In {\em {Proc. CVPRW}}, June 2020.

\bibitem{kendall2017end}
Alex Kendall, Hayk Martirosyan, Saumitro Dasgupta, Peter Henry, Ryan Kennedy,
  Abraham Bachrach, and Adam Bry.
\newblock End-to-end learning of geometry and context for deep stereo
  regression.
\newblock In {\em {Proc. CVPR}}, pages 66--75, 2017.

\bibitem{khamis2018stereonet}
Sameh Khamis, Sean Fanello, Christoph Rhemann, Adarsh Kowdle, Julien Valentin,
  and Shahram Izadi.
\newblock Stereonet: Guided hierarchical refinement for real-time edge-aware
  depth prediction.
\newblock In {\em {Proc. ECCV}}, pages 573--590, 2018.

\bibitem{adam}
Diederik~P Kingma and Jimmy Ba.
\newblock Adam: A method for stochastic optimization.
\newblock {\em arXiv preprint arXiv:1412.6980}, 2014.

\bibitem{klaus2006segment}
Andreas Klaus, Mario Sormann, and Konrad Karner.
\newblock Segment-based stereo matching using belief propagation and a
  self-adapting dissimilarity measure.
\newblock In {\em {Proc. ICPR}}, volume~3, pages 15--18, 2006.

\bibitem{li2020revisiting}
Zhaoshuo Li, Xingtong Liu, Nathan Drenkow, Andy Ding, Francis~X Creighton,
  Russell~H Taylor, and Mathias Unberath.
\newblock Revisiting stereo depth estimation from a sequence-to-sequence
  perspective with transformers.
\newblock {\em arXiv preprint arXiv:2011.02910}, 2020.

\bibitem{liang2018learning}
Zhengfa Liang, Yiliu Feng, Yulan Guo, Hengzhu Liu, Wei Chen, Linbo Qiao, Li
  Zhou, and Jianfeng Zhang.
\newblock Learning for disparity estimation through feature constancy.
\newblock In {\em {Proc. CVPR}}, pages 2811--2820, 2018.

\bibitem{lipson2021raft}
Lahav Lipson, Zachary Teed, and Jia Deng.
\newblock Raft-stereo: Multilevel recurrent field transforms for stereo
  matching.
\newblock {\em arXiv preprint arXiv:2109.07547}, 2021.

\bibitem{loop1999computing}
Charles Loop and Zhengyou Zhang.
\newblock Computing rectifying homographies for stereo vision.
\newblock In {\em Proceedings. 1999 IEEE Computer Society Conference on
  Computer Vision and Pattern Recognition (Cat. No PR00149)}, volume~1, pages
  125--131. IEEE, 1999.

\bibitem{luo2020wavelet}
Chenchi Luo, Yingmao Li, Kaimo Lin, George Chen, Seok-Jun Lee, Jihwan Choi,
  Youngjun~Francis Yoo, and Michael~O Polley.
\newblock Wavelet synthesis net for disparity estimation to synthesize dslr
  calibre bokeh effect on smartphones.
\newblock In {\em {Proc. CVPR}}, pages 2407--2415, 2020.

\bibitem{mayer2018makes}
Nikolaus Mayer, Eddy Ilg, Philipp Fischer, Caner Hazirbas, Daniel Cremers,
  Alexey Dosovitskiy, and Thomas Brox.
\newblock What makes good synthetic training data for learning disparity and
  optical flow estimation?
\newblock {\em {International Journal of Computer Vision}}, 126(9):942--960,
  2018.

\bibitem{mayer2016large}
Nikolaus Mayer, Eddy Ilg, Philip Hausser, Philipp Fischer, Daniel Cremers,
  Alexey Dosovitskiy, and Thomas Brox.
\newblock A large dataset to train convolutional networks for disparity,
  optical flow, and scene flow estimation.
\newblock In {\em {Proc. CVPR}}, pages 4040--4048, 2016.

\bibitem{menze2015object}
Moritz Menze and Andreas Geiger.
\newblock Object scene flow for autonomous vehicles.
\newblock In {\em {Proc. CVPR}}, pages 3061--3070, 2015.

\bibitem{pang2017cascade}
Jiahao Pang, Wenxiu Sun, Jimmy~SJ Ren, Chengxi Yang, and Qiong Yan.
\newblock Cascade residual learning: A two-stage convolutional neural network
  for stereo matching.
\newblock In {\em {Proc. CVPRW}}, pages 887--895, 2017.

\bibitem{pang2018zoom}
Jiahao Pang, Wenxiu Sun, Chengxi Yang, Jimmy Ren, Ruichao Xiao, Jin Zeng, and
  Liang Lin.
\newblock Zoom and learn: Generalizing deep stereo matching to novel domains.
\newblock In {\em {Proc. CVPR}}, pages 2070--2079, 2018.

\bibitem{pytorch}
Adam Paszke, Sam Gross, Francisco Massa, Adam Lerer, James Bradbury, Gregory
  Chanan, Trevor Killeen, Zeming Lin, Natalia Gimelshein, Luca Antiga, Alban
  Desmaison, Andreas Kopf, Edward Yang, Zachary DeVito, Martin Raison, Alykhan
  Tejani, Sasank Chilamkurthy, Benoit Steiner, Lu Fang, Junjie Bai, and Soumith
  Chintala.
\newblock Pytorch: An imperative style, high-performance deep learning library.
\newblock In H. Wallach, H. Larochelle, A. Beygelzimer, F. d\textquotesingle
  Alch\'{e}-Buc, E. Fox, and R. Garnett, editors, {\em Advances in Neural
  Information Processing Systems 32}, pages 8024--8035. Curran Associates,
  Inc., 2019.

\bibitem{bokeh}
Lars Rehm.
\newblock Evaluating computational bokeh: How we test smartphone portrait
  modes.
\newblock
  \url{https://www.dxomark.com/evaluating-computational-bokeh-test-smartphone-portrait-modes/},
  2018.
\newblock Accessed: 2021-11-15.

\bibitem{scharstein2014high}
Daniel Scharstein, Heiko Hirschm{\"u}ller, York Kitajima, Greg Krathwohl, Nera
  Ne{\v{s}}i{\'c}, Xi Wang, and Porter Westling.
\newblock High-resolution stereo datasets with subpixel-accurate ground truth.
\newblock In {\em German Conference on Pattern Recognition}, pages 31--42,
  2014.

\bibitem{scharstein2002taxonomy}
Daniel Scharstein and Richard Szeliski.
\newblock A taxonomy and evaluation of dense two-frame stereo correspondence
  algorithms.
\newblock {\em {International Journal of Computer Vision}}, 47(1):7--42, 2002.

\bibitem{middlebury}
Daniel Scharstein and Richard Szeliski.
\newblock A taxonomy and evaluation of dense two-frame stereo correspondence
  algorithms.
\newblock {\em {International Journal of Computer Vision}}, 47(1):7--42, 2002.

\bibitem{eth3d}
Thomas Schops, Johannes~L Schonberger, Silvano Galliani, Torsten Sattler,
  Konrad Schindler, Marc Pollefeys, and Andreas Geiger.
\newblock A multi-view stereo benchmark with high-resolution images and
  multi-camera videos.
\newblock In {\em {Proc. CVPR}}, pages 3260--3269, 2017.

\bibitem{airsim2017fsr}
Shital Shah, Debadeepta Dey, Chris Lovett, and Ashish Kapoor.
\newblock Airsim: High-fidelity visual and physical simulation for autonomous
  vehicles.
\newblock In {\em Field and Service Robotics}, 2017.

\bibitem{shen2021cfnet}
Zhelun Shen, Yuchao Dai, and Zhibo Rao.
\newblock Cfnet: Cascade and fused cost volume for robust stereo matching.
\newblock In {\em {Proc. CVPR}}, pages 13906--13915, 2021.

\bibitem{song2021adastereo}
Xiao Song, Guorun Yang, Xinge Zhu, Hui Zhou, Zhe Wang, and Jianping Shi.
\newblock Adastereo: a simple and efficient approach for adaptive stereo
  matching.
\newblock In {\em {Proc. CVPR}}, pages 10328--10337, 2021.

\bibitem{autoflow}
Deqing Sun, Daniel Vlasic, Charles Herrmann, Varun Jampani, Michael Krainin,
  Huiwen Chang, Ramin Zabih, William~T Freeman, and Ce Liu.
\newblock Autoflow: Learning a better training set for optical flow.
\newblock In {\em {Proc. CVPR}}, pages 10093--10102, 2021.

\bibitem{sun2021loftr}
Jiaming Sun, Zehong Shen, Yuang Wang, Hujun Bao, and Xiaowei Zhou.
\newblock Loftr: Detector-free local feature matching with transformers.
\newblock In {\em {Proc. CVPR}}, pages 8922--8931, 2021.

\bibitem{sun2003stereo}
Jian Sun, Nan-Ning Zheng, and Heung-Yeung Shum.
\newblock Stereo matching using belief propagation.
\newblock {\em {IEEE Trans. on Pattern Analysis and Machine Intelligence}},
  25(7):787--800, 2003.

\bibitem{taniai2017continuous}
Tatsunori Taniai, Yasuyuki Matsushita, Yoichi Sato, and Takeshi Naemura.
\newblock Continuous 3d label stereo matching using local expansion moves.
\newblock {\em {IEEE Trans. on Pattern Analysis and Machine Intelligence}},
  40(11):2725--2739, 2017.

\bibitem{tankovich2021hitnet}
Vladimir Tankovich, Christian Hane, Yinda Zhang, Adarsh Kowdle, Sean Fanello,
  and Sofien Bouaziz.
\newblock Hitnet: Hierarchical iterative tile refinement network for real-time
  stereo matching.
\newblock In {\em {Proc. CVPR}}, pages 14362--14372, 2021.

\bibitem{teed2020raft}
Zachary Teed and Jia Deng.
\newblock Raft: Recurrent all-pairs field transforms for optical flow.
\newblock In {\em {Proc. ECCV}}, pages 402--419, 2020.

\bibitem{fallingthings}
Jonathan Tremblay, Thang To, and Stan Birchfield.
\newblock Falling things: A synthetic dataset for 3d object detection and pose
  estimation.
\newblock In {\em {Proc. CVPRW}}, pages 2038--2041, 2018.

\bibitem{van2002hierarchical}
Geert Van~Meerbergen, Maarten Vergauwen, Marc Pollefeys, and Luc Van~Gool.
\newblock A hierarchical symmetric stereo algorithm using dynamic programming.
\newblock {\em {International Journal of Computer Vision}}, 47(1):275--285,
  2002.

\bibitem{wang2020improving}
Jialiang Wang, Varun Jampani, Deqing Sun, Charles Loop, Stan Birchfield, and
  Jan Kautz.
\newblock Improving deep stereo network generalization with geometric priors.
\newblock {\em arXiv preprint arXiv:2008.11098}, 2020.

\bibitem{xu2020aanet}
Haofei Xu and Juyong Zhang.
\newblock Aanet: Adaptive aggregation network for efficient stereo matching.
\newblock In {\em {Proc. CVPR}}, pages 1959--1968, 2020.

\bibitem{yan2019segment}
Tingman Yan, Yangzhou Gan, Zeyang Xia, and Qunfei Zhao.
\newblock Segment-based disparity refinement with occlusion handling for stereo
  matching.
\newblock {\em {IEEE Trans. on Image Processing}}, 28(8):3885--3897, 2019.

\bibitem{yang2019hierarchical}
Gengshan Yang, Joshua Manela, Michael Happold, and Deva Ramanan.
\newblock Hierarchical deep stereo matching on high-resolution images.
\newblock In {\em {Proc. CVPR}}, pages 5515--5524, 2019.

\bibitem{yang2008stereo}
Qingxiong Yang, Liang Wang, Ruigang Yang, Henrik Stew{\'e}nius, and David
  Nist{\'e}r.
\newblock Stereo matching with color-weighted correlation, hierarchical belief
  propagation, and occlusion handling.
\newblock {\em {IEEE Trans. on Pattern Analysis and Machine Intelligence}},
  31(3):492--504, 2008.

\bibitem{yu2017dilated}
Fisher Yu, Vladlen Koltun, and Thomas Funkhouser.
\newblock Dilated residual networks.
\newblock In {\em {Proc. CVPR}}, pages 472--480, 2017.

\bibitem{zbontar2015computing}
Jure Zbontar and Yann LeCun.
\newblock Computing the stereo matching cost with a convolutional neural
  network.
\newblock In {\em {Proc. CVPR}}, pages 1592--1599, 2015.

\bibitem{zhang2019ga}
Feihu Zhang, Victor Prisacariu, Ruigang Yang, and Philip~HS Torr.
\newblock Ga-net: Guided aggregation net for end-to-end stereo matching.
\newblock In {\em {Proc. CVPR}}, pages 185--194, 2019.

\bibitem{zhang1998determining}
Zhengyou Zhang.
\newblock Determining the epipolar geometry and its uncertainty: A review.
\newblock {\em International journal of computer vision}, 27(2):161--195, 1998.

\bibitem{zhu2019deformable}
Xizhou Zhu, Han Hu, Stephen Lin, and Jifeng Dai.
\newblock Deformable convnets v2: More deformable, better results.
\newblock In {\em {Proc. CVPR}}, pages 9308--9316, 2019.

\end{thebibliography}
}

\end{document}